\documentclass[runningheads]{llncs}

\usepackage{eccv}

\usepackage{eccvabbrv}

\usepackage{graphicx}
\usepackage{booktabs}
\usepackage{subcaption}
\usepackage{amsmath,amssymb,amsfonts}
\usepackage{algorithmic}
\usepackage{adjustbox}
\usepackage{graphics}
\usepackage{textcomp}
\usepackage{multirow}
\usepackage{multicol}
\usepackage{xurl}
\usepackage{color, soul}
\usepackage{tabularx}
\usepackage{changepage}
\usepackage{rotating}
\usepackage{array}

\usepackage[accsupp]{axessibility}  

\usepackage{hyperref}

\usepackage{orcidlink}

\begin{document}

\title{Autobiasing Event Cameras} 


\author{Mehdi Sefidgar Dilmaghani\inst{1}\orcidlink{0000-0003-1975-7815} \and
Waseem Shariff\inst{1}\orcidlink{0000-0001-7298-9389} \and
Cian Ryan\inst{2}\orcidlink{0000-0002-7353-6936} \and
Joseph Lemley\inst{2}\orcidlink{0000-0002-0595-2313} \and
Peter Corcoran\inst{1}\orcidlink{0000-0003-1670-4793}} 

\authorrunning{M.~Sefidgar Dilmaghani et al.}

\institute{University of Galway, University Road, Galway, Ireland\\
\email{\{M.SefidgarDilmaghani1, W.Shariff1, Peter.Corcoran\}@universityofgalway.ie}\and
Tobii, Business Park Block 5, Brockagh, Parkmore East, Galway, Ireland\\
\email{\{Cian.Ryan, Joseph.Lemley\}@tobii.com}}

\maketitle

\begin{abstract}
  This paper presents an autonomous method to address challenges arising from severe lighting conditions in machine vision applications that use event cameras. To manage these conditions, the research explores the built-in potential of these cameras to adjust pixel functionality, named bias settings. As cars are driven at various times and locations, shifts in lighting conditions are unavoidable. Consequently, this paper utilizes the neuromorphic YOLO-based face tracking module of a driver monitoring system as the event-based application to study. The proposed method uses numerical metrics to continuously monitor the performance of the event-based application in real-time. When the application malfunctions, the system detects this through a drop in the metrics and automatically adjusts the event camera's bias values. The Nelder-Mead simplex algorithm is employed to optimize this adjustment, with fine-tuning continuing until performance returns to a satisfactory level. The advantage of bias optimization lies in its ability to handle conditions such as flickering or darkness without requiring additional hardware or software. To demonstrate the capabilities of the proposed system, it was tested under conditions where detecting human faces with default bias values was impossible. These severe conditions were simulated using dim ambient light and various flickering frequencies. Following the automatic and dynamic process of bias modification, the metrics for face detection significantly improved under all conditions. Autobiasing resulted in an increase in the YOLO confidence indicators by more than 33\% for object detection and 37\% for face detection highlighting the effectiveness of the proposed method.
  \keywords{Autobiasing \and Bias \and Event Cameras \and Neuromorphic Sensors \and Optimization}
\end{abstract}

\section{Introduction}
\label{sec:intro}
Event cameras have emerged as a reliable solution for computer vision applications, primarily due to their unique approach to scene capture. Unlike frame-based cameras, these sensors record changes in light and movement in the environment by adjusting the values of individual pixels independently \cite{m1}. When the sensor detects a change in light intensity due to movement, it generates an event that includes the X and Y positions of the pixel, the timestamp of occurrence (in microsecond), and the polarity of the change which can be positive or negative depending on the change in the light. This variable data captured by event cameras effectively mitigates the issue of under-sampling encountered in conventional frame-based cameras \cite{w25}. This feature is particularly beneficial in automotive technologies where precise and detailed information is essential. Event cameras are also noted for their low latency and high temporal resolution, which enhances their suitability for various automotive applications, including out-of-cabin object detection and in-cabin driver monitoring systems \cite{w25}. Recent years have witnessed a growing interest among researchers in event-based driver monitoring systems (DMS), leading to the development of several initiatives \cite{m3, w25, w26}.

In the design of DMS, it is crucial to consider that vehicles operate under diverse lighting conditions, ranging from daylight to nighttime, and across varying weather conditions, such as sunny or cloudy skies, and in different environments like open roads or tunnels. In addition to these variations in lighting, modern vehicles contain numerous sources of noise and flickering, including large screens and various sensors, which can negatively impact the functionality of any type of camera. The inherent sensitivity of event cameras to changes in ambient light can present challenges when implementing them in DMS applications. However, event cameras offer an advantage in addressing lighting challenges due to their flexibility in allowing users to adjust sensor-level settings to adapt to ambient lighting conditions. This flexibility, provided by event camera manufacturers, is referred to as bias \cite{m4}.

By adjusting the bias settings, it becomes possible to tune the characteristics of the internal circuits of the sensor, particularly different filters, to ensure compatibility with various conditions. These settings include the camera's sensitivity to positive or negative light changes, determining the minimum light change required to trigger an event. Additionally, two other biases control the sharpness of a low-pass filter, which removes background noise, and a high-pass filter, responsible for filtering out low-speed movements from the event stream. Lastly, another bias that determines the sleeping time for each pixel after each event generation. More details about the event camera hardware and these biases are presented in section 2 \cite{m4}.

This study investigates the use of bias settings in event cameras to enhance the adaptability of DMS applications with diverse lighting conditions. Specifically, the research aims to explore the implementation of such adaptation mechanisms, including when and how the bias settings should be adjusted, the extent of the required changes, and the specific biases that need modification. Recognizing the impracticality of manual adjustments by drivers and passengers, the study seeks to develop an automated system capable of monitoring camera performance across various lighting conditions. This system would dynamically adjust the bias settings in response to changing conditions, ensuring optimal camera functionality without requiring manual intervention.

In this study, the term "autobiasing" refers to the automatic and dynamic adjustment of biases within event cameras. Previous research has explored autobiasing but focused primarily on single bias adjustments, such as Nair et al.'s feedback control loop that regulates the refractory period bias to maintain event count within a desired range \cite{m11}. In contrast, this study introduces a novel approach by simultaneously tuning all biases, thus providing a more versatile solution applicable to various event-based applications. Unlike previous systems, which were designed to address specific issues and incorporated separate monitoring blocks, the proposed algorithm continuously adapts to any bias-related changes and optimizes performance dynamically. Additional research by Dilmaghani et al. has demonstrated the impact of bias changes on event camera output sharpness and DMS functionality \cite{m5, m6}, while another study investigated auto-focus capabilities \cite{m7}. This study's significance lies in its ability to ensure optimal performance across diverse lighting and noise conditions, addressing a critical challenge in event-based systems and enhancing the reliability of event cameras compared to traditional camera types.

According to \cite{m6}, face detection and tracking are critical for the effective operation of Driver Monitoring Systems (DMS), as they underpin other components such as eye tracking and blink counting. This project aims to continuously monitor face tracking metrics to evaluate the DMS's face detection capabilities. This study utilizes a YOLO V3-based \cite{m18} network for face detection and assess the proposed bias optimization method based on YOLO's output confidences. It is important to note that no retraining, fine tuning, or evaluation of YOLO was done in this study as such efforts would be out of scope. Dynamic bias adjustment will be implemented by feeding monitoring results into a bias controller. If DMS functionality metrics fall below a specified threshold, the simplex algorithm will optimize the bias values until performance metrics exceed the threshold. The proposed algorithm includes the following main improvements on the state of the art:
\begin{itemize}
    \item Represents the first known approach to optimize all biases in an event camera system simultaneously. This method effectively addresses issues related to adverse lighting conditions, such as flickering and darkness, significantly improving performance.
    \item Continuously monitors the performance of DMS components using quantitative metrics, focusing particularly on face tracking capabilities. This real-time tracking allows for immediate detection and correction of any performance declines.
    \item Evaluates the effectiveness of real-time autobiasing in enhancing DMS functionality. Employing YOLO V3 for face detection and a simplex algorithm for bias adjustments, the study demonstrates significant improvements in system performance and reliability.
\end{itemize}

The paper is organized as follows: the process of event generation and the impact of biases are described in section 2. A summary of the state of the art is given in Section 3. In Section 4, the problem addressed in this study and the approach taken to solve it are explained in extensive detail. The algorithm test results are proposed in Section 5, and the conclusion is covered in Section 6.

\section{Event Camera Hardware and Biases}
\begin{figure}[h]
  \centering
  \includegraphics[width = 10 cm]{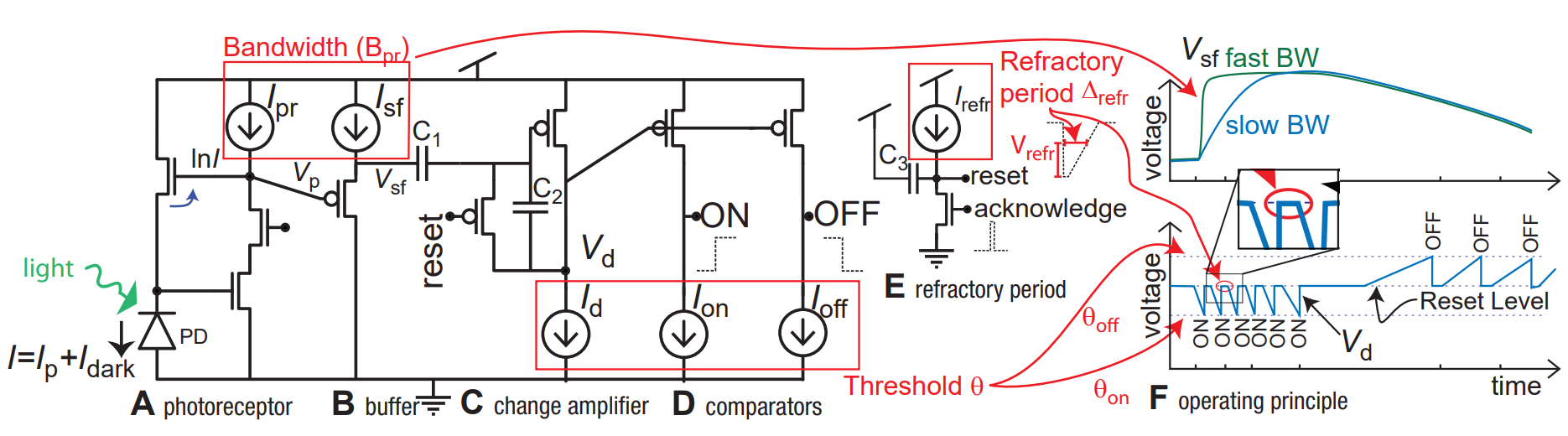}
  \caption{The schematic of each pixel in event cameras \cite{w19}.}
  \label{fig:0}
\end{figure}

Understanding how the event cameras work and the part and effect that biasing plays in this process is crucial before delving too deeply into the proposed autobiasing system. The schematic of a single event camera pixel is displayed in Figure \ref{fig:0}. In stage A, the photodiode (PD) senses changes in light levels in the environment and transforms them into electrical current. In step B, the current is filtered and buffered. Together, Ip and Ipr define the low pass filter's first stage cut-off frequency, which determines the bandwidth Bpr. The second stage cut-off frequency is determined by Isf, which is governed by bias\_fo. The flickering and low frequency background noise can be removed by adjusting the bias\_fo. However, as the graphs in figure \ref{fig:0} show, this modification results in a change in the sensors' operating speed. Therefore, there is a trade-off between the sensor's speed and level of noise mitigation. In subsequent stages, the ratio between C1 and C2, along with the values of Ion, Ioff, and Id, dictate the sensor's sensitivity to light changes. As previously stated, event cameras capture environmental light changes and produce events indicating whether the change was positive or negative. This stage controls the thresholds that must be exceeded for a positive or negative event to be generated, with bias\_diff\_on and bias\_diff\_off responsible for adjusting the thresholds. The voltage and current sources, referred to as Vref and Iref, along with capacitor C3, determine the refractory period duration. This period represents the interval after each event generation during which the sensor remains inactive until the next event is generated. The bias responsible for setting the refractory period duration is termed bias\_refr. Prophesee cameras feature an additional bias known as bias\_hpf, which adjusts a high-pass filter to determine the minimum speed required for a moving object to be detected by the sensor.

The above process can be better understood with the aid of the graphs on the right side of the figure \ref{fig:0}. The output voltage is stable when there are no variations in the ambient light level; when the changes happen, the voltage begins to rise or fall. The variation continues until the voltage level goes beyond the thresholds, either positive or negative, at which point an event is generated. After that, the voltage level is reset to its initial, stable value, remains there, and doesn't record any changes to the event until the refractory time is over. The output voltage changes again until the next event is generated.

Table \ref{tbl:0} lists the biases included in the Prophesee IMX 636 camera, detailing the consequence of any adjustments. 

\newcolumntype{C}[1]{>{\centering\arraybackslash}m{#1}}

\begin{table}[h]
    \centering
    \fontsize{7pt}{9pt}\selectfont
    \caption{Introduction and effects of increasing bias settings in the Prophesee IMX 636 event camera \cite{w13}.}
    
    \begin{tabular}{|C{1 cm}|C{8.5 cm}|C{2.3 cm}|}
    \hline
    \textbf{Bias} & \textbf{Description} & \textbf{Bias Increment}  \\
    \hline
    diff\_on \& off & Set the positive and negative contrast thresholds, determining the minimum change in light intensity required to generate an event. & Higher contrast threshold \\
    \hline
    fo & Adjusts the low-pass filter bandwidth to remove flickering noise at the expense of decreased sensor speed. & Wider bandwidth \\
    \hline
    hpf & Adjusts the high-pass filter to filter out slow-changing elements from the scene, thereby showing only fast-moving objects.  & Sensitive to faster light changes \\
    \hline
    refr & Sets the sleep duration required by the sensor after generating each event. & Shorter slept period \\
    \hline

    \end{tabular}
    
    \label{tbl:0}
\end{table}

\section{Literature Review}
In this section we explore literature on framed-based cameras and auto-exposure, followed by bias tuning of event cameras.

\subsubsection{Conventional Camera Auto-exposure}
The following aggregates findings from a range of studies that address the challenges of auto exposure and noise reduction in conventional frame-based cameras, emphasizing the role of hardware innovations in improving image quality. 

Gomez et al. \cite{w1} introduce an algorithm for fusing multi exposure and multi focus images to counteract camera shake, highlighting a significant step forward in dynamic range and focus enhancement. Skorka et al. \cite{w2} explore the IEEE P2020 Noise standard, detailing methods for evaluating noise at both sensor and camera levels, with a particular focus on automotive applications. Liu Jia-wen \cite{w3} reviews the evolution of auto-focus methods, illustrating the importance of hardware in achieving sharp images. Mallick et al. \cite{w4} characterize the noise in Kinect depth images and suggest hardware configurations for noise reduction. Gao et al. \cite{w5} investigate acoustic meta materials, offering insights that, although focused on sound, parallel the challenges in visual noise reduction.

Kaur et al. \cite{w6} and Al Mudhafar et al. \cite{w7} discuss image enhancement and noise removal strategies, pointing to the crucial role of hardware adjustments. Prasad et al. \cite{w8} provide a systematic overview of noise types, denoising methods, and the importance of hardware settings in noise management. Konnik et al. \cite{w9} offer a model of CCD and CMOS photo sensors, focusing on reducing hardware-induced noise. Finally, Koli et al. \cite{w10} review impulse noise reduction techniques, emphasizing hardware's effectiveness in minimizing noise without compromising image integrity. Collectively, these studies highlight the important role of hardware advancements in enhancing auto exposure and reducing noise, offering valuable insights into the continuous improvement of camera technologies.

\subsubsection{Event Camera Bias Setting}
An essential aspect of optimizing the performance of EC (or Dynamic Vision Sensor (DVS)) is the proper setting of bias parameters, which directly affect the camera's sensitivity, noise performance, and overall output quality. This sub-section of literature review focuses on the advancements and methodologies related to the bias setting in event cameras, aiming to enhance the quality of their outputs.

Authors in \cite{w11} laid the foundational work in the field of neuromorphic vision sensors with their development of a 128×128 pixel event camera, highlighting the importance of asynchronous temporal contrast vision sensors in achieving high dynamic range and low latency. This work underscored the necessity of understanding and optimizing bias settings to maximize sensor performance. Further, detailed guides and documentation, such as those provided by \cite{w12} and \cite{w13}, offer valuable insights into biasing dynamic sensors, emphasizing the qualitative aspects of bias adjustment and its impact on sensor performance. These resources, however, often lack a comprehensive discussion on the trade-offs involved in bias adjustments, highlighting a gap in quantitative analysis within the field.

The impact of bias settings on key performance metrics under varied illumination levels was explored in the works of \cite{w14}, \cite{w15}, \cite{w16}, which measured and reported DVS performance, albeit without a full characterization across different bias settings. Graca et al. \cite{w17}, \cite{w18} and Delbruck et al.  \cite{w19} furthered this research by examining how DVS noise performance and bandwidth are influenced by illumination and photoreceptor bias, presenting an in-depth analysis of the optimal biasing and physical limits of DVS event noise. Delbruck et al. \cite{w19} introduced a novel approach by implementing a feedback control algorithm to dynamically tune bias settings, aiming to maintain the output event rate within a target window. This method represents a significant step toward adaptive bias management, potentially enhancing the robustness and efficiency of event camera operation. 

Moreover, the effect of biasing on the output sharpness of event cameras was investigated by Dilmaghani et al. \cite{m5}, providing valuable insights into controlling and evaluating output quality through bias adjustments. McReynolds et al. \cite{w21} proposed an experimental method to predict the performance of event cameras, emphasizing the importance of understanding the relationship between bias settings and inferred parameters such as Temporal Contrast event threshold, pixel bandwidth, and refractory period. This work is instrumental in developing realistic DVS models and simulators, as highlighted by the efforts of  \cite{w22}, \cite{w23}, \cite{w24}, who have contributed to the creation of open event camera simulators and methodologies to translate video frames to realistic DVS events.

\section{Methodology}
Event cameras capture light changes at the pixel level, offering faster and more detailed data recording ideal for high temporal resolution applications. However, despite these advantages, event cameras face significant challenges, particularly under variable lighting conditions. Driver monitoring systems are an example of applications where event camera capabilities are extremely useful, yet shifts in lighting conditions during driving are unavoidable. Low light, flashing lights, and noisy environments can degrade output quality and complicate accurate in-cabin monitoring tasks. Effectively addressing these challenges is crucial for enhancing their potential, whether used alone or besides other cameras in DMS applications \cite{m3}.

The primary objective of this research is to address the above-mentioned challenges in real time and at low computational cost, with the aim of exploiting the unique features of time event cameras. As mentioned previously, the bias settings in the event camera are designed to control the event stream. The bias\_fo, for instance, can be used to reduce the low frequency background noise from the output stream of the cameras \cite{m4}. However, because event cameras function in a variety of environments, fine-tuning the biases at each new condition is not a straightforward operation.

An algorithm that continually monitors the performance of event-based applications and in the case that an application malfunctions, automatically adjusts the biases until an optimal performance is achieved, is the best possible approach. The focus of this research is to design and implement such an algorithm. This research utilizes the modified YOLO-V3 described in \cite{m3} for face detection to test and demonstrate the effectiveness of the proposed algorithm to improve the effectiveness of a pre-existing time event based neural network. The algorithm monitors the network's performance based on a parameter that will be described in the next sections. When the system detects a decline in performance, it automatically adjusts all of the event cameras' biases until the face detection performance meets the predetermined criteria. The next sections provide an explanation of the autobiasing pipeline and each block in the pipeline.

\subsection{Autobiasing Pipeline}
\begin{figure}[h]
    \centering
    \begin{minipage}{0.45\textwidth} 
        \centering
        \includegraphics[height=2.5cm]{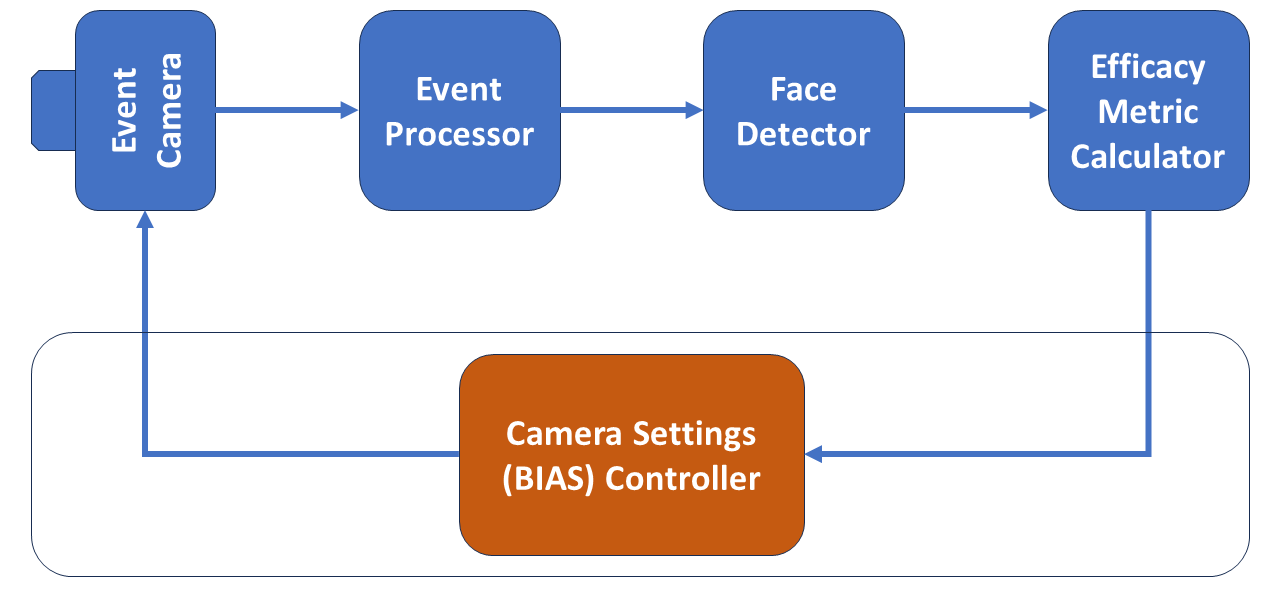} 
        \caption{The block diagram of the proposed autobiasing system.}
        \label{fig:1}
    \end{minipage}
    \hspace{1cm} 
    \begin{minipage}{0.45\textwidth} 
        \centering
        \includegraphics[height=2.5 cm]{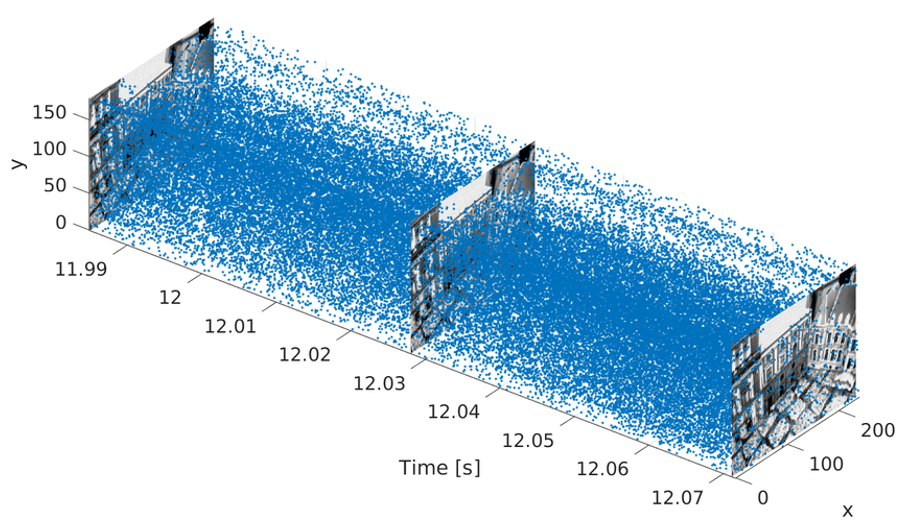} 
        \caption{Accumulation of event stream data in 2D frame matrices over time.\cite{m9}}
        \label{fig:2}
    \end{minipage}
\end{figure}

Figure \ref{fig:1} depicts the block diagram of the autobiasing system. The event processor first processes the raw events, that the camera captures, in order to format them correctly based on the requirements of the application, which is a YOLO-V3 based face tracking network here. These restructured events are then ready for the application to process. In the efficacy metric calculator the application's performance is evaluated and monitored. The results of this assessment will then be fed to the bias controller, enabling it to adjust the camera's settings such that the application achieves optimal performance, for the subsequent events. The next section offers additional details on each of the blocks.

\subsection{Building Blocks}

\subsubsection{Event Camera}
The event camera utilized in this research is a Metavision EVK4-HD, with a IMX-636 sensor manufactured by Sony and Prophesee. The latency of the camera in the 1k lux is less than 220 us, the maximum bandwidth is 1.6 Gbps, and the maximum system power via USB is 1.5W as stated in the manufacturers website \cite{m8}.

\subsubsection{Events Processor}
Even though this block is not a key element of the autobiasing process, understanding exactly what it is helps to comprehend the algorithm. The face detection algorithm employed in this study is a modified version of the YOLO-V3, and it receives the frames as input, similar to many other computer vision applications. Since the event cameras output is stream of events, this block is required to receive the raw event stream and convert it to frames. 

The camera generates an event whenever there is a change in the ambient light. Events are composed of a timestamp (in microseconds), the x and y coordinates where the event occurred, and a polarity. The polarity is a binary indication of the change in light intensity at the specific location and time. A stream of events is a series of events that occur one after another across time. In order to transform this stream into frames, it is necessary to collect batches of events within predetermined time frames or event counts. Frame-by-frame accumulation of events is seen in Figure \ref{fig:2}.

This task's algorithm is developed such that it is possible to specify the number of frames produced each second as well as the duration of time that each frame's events accumulate. The frame rate of 8 frames per second was selected for this project. Lower frame rates result in information loss, while higher rates impose excessive computational burden on the processor, leading to delays that jeopardize real-time operations.

It is crucial to remember that in the autobiasing process, application input has no impact on the main mechanism. The autobiasing procedure monitors the output and performance of the application, as opposed to concentrating on its input side. Hence, autobiasing is not constrained by the input format being frames and can effectively operate for applications beyond face detection. It remains functional even with diverse input formats, like raw event streams, provided there is a method to assess the application's functionality efficiency.

\subsubsection{Face Detector}
The main objective of autobiasing is to maximize an event-based application's performance. This study proposes a method that uses quantitative measurements of face tracking to monitor an application's performance and adjusts the event cameras' settings to enhance those metric.

To demonstrate the advantages of autobiasing, this study utilizes a face detection algorithm based on YOLO-V3 architecture, which serves as an example of an event-based application adopted from \cite{m3}. The innovative design, called gated recurrent YOLO (GR-YOLO), is used for multi-object tracking and recognition, with a particular emphasis on faces and eyes. To address the temporal sparsity of events, GR-YOLO combines a fully convolutional gated recurrent unit (GRU) layer. A manually annotated dataset from a Prophesee event camera and a synthetic dataset called N-Helen, which was created using pre-existing RGB datasets with facial landmark annotations are used to train the network . Then the network is qualitatively evaluated on actual driving event data \cite{m3}.

This detector's main goal is to recognize faces and eyes in an accurate and efficient manner. The bounding box center (bx and by), width and height (bw and bh), the probability of an object being identified (to), and class probabilities (p1 and p2) indicating the likelihood of belonging to each class (face and eye) are all included in each prediction that the model outputs. Following the prediction stage, the model employs post-processing methods to fine-tune and choose the most accurate and relevant bounding box predictions, such as the probability of object presence or the probability of detecting a face \cite{m3}.

\subsubsection{Efficacy Metric Calculator}
This block is application-specific and plays a key role in the autobiasing algorithm by determining the starting moments of bias adjustments. If the metric falls below a predetermined lower threshold, autobiasing initiates. Following dynamic adjustments to the biases, autobiasing ceases when the it has effectively optimized the application's efficiency. The stopping moment of the autobiasing will be explained in the next section.

The metric employed in this research is based on the face tracking metric for the event-based driver monitoring system proposed in \cite{m6}. This metric is referred to as efficacy metric, and it compares the number of frames with detected faces to the total number of frames in a certain time period, which in this study is one second. It is important to highlight that adjusting this value does not impact the overall functionality of the system; rather, it reflects the specific design of the current system. Here's how the face detection performance metric is calculated:

\begin{equation}
  Efficacy Metric = \frac{F_{detected}}{F_{total}}
\end{equation}

$F_{detected}$ represents the number of frames with a detected face, while $F_{total}$ represents the overall number of frames in a second. The lower threshold at which the autobiasing starts is when the efficacy metric goes below 0.5. It is crucial to emphasize that the project requirements and user decisions play a major role in determining the threshold value and time frame. Changing these settings doesn't affect the system's functionality or structure, and has no technical impact.

\subsubsection{Bias Controller}
The heart of the autobiasing is this block. When an application's functionality drops below an established minimum level, this controller begins adjusting the bias values in a way that the application's functionality rises back to the desired level. So, the objective of this block is to address a specific issue: the event camera records facial data, which is then processed by the face detector to detect faces. However, in cases of poor lighting conditions, the face detection function fails. Therefore, it becomes essential to optimize the bias values of the event camera to produce frames in which faces can be accurately detected. The bias values then undergo an optimization process, which is explained below. Subsequently, the newly optimized bias values are applied to the camera settings. Following this adjustment, the application's functionality is reassessed after a 2-second interval to determine whether the bias optimization process should continue or stop. The criteria of stopping the optimization process is based on the changes required in the bias value. In this project, when the required change for each bias is less than 1, the process stops.

Table \ref{tbl:methods} provides a brief explanation of potential methods considered for the proposed multi-input single-output black box optimization problem. Based on this table, the Nelder-Mead Simplex method \cite{m10} is chosen as the primary optimization algorithm for the autobiasing system.


\begin{table}[h]
    \centering
    \fontsize{7pt}{9pt}\selectfont
    \caption{Comparison of optimization methods for event camera bias settings}
    
    \begin{tabular}{|C{2.9 cm}|C{9 cm}|}
    \hline
    \textbf{Method}  & \textbf{Summary of Characteristics} \\
    \hline
    Genetic Algorithms \cite{m12} &
    Versatile, applicable to broad problems, handles complex landscapes; computationally expensive with large populations. \\
    \hline
    CMA-ES \cite{m13} &
    Effective on continuous, rugged landscapes; requires multiple function evaluations, potentially costly computationally. \\
    \hline
    PSO \cite{m14} &
    Fast convergence in some scenarios, comparable computational cost to GA; requires managing a population of solutions. \\
    \hline
    Simulated Annealing \cite{m15}&
    Less computationally expensive than population-based methods, effective for escaping local optima; sensitive to annealing schedule. \\
    \hline
    Bayesian Optimization \cite{m16} &
    Efficient in objective function evaluations, intelligently selects points; computationally intensive due to Gaussian Process modeling. \\
    \hline
    Random Search \cite{m17} &
    Simple, low computational complexity; may require many evaluations, lacks systematic exploration. \\
    \hline
    Nelder-Mead Simplex \cite{m10} &
    No derivatives needed, suitable for black-box optimization; straightforward, effective for lower-dimensional problems; less computationally expensive than population-based methods. \\
    \hline    
    
    \end{tabular}
    
    \label{tbl:methods}
\end{table}

 For nonlinear optimization problems, the Nelder-Mead simplex algorithm, also referred to as the downhill simplex method, is an iterative optimization strategy. The method modifies a geometric shape called a simplex. This shape consists of points in the parameter space, representing the bias values obtained from the camera in this study. The modification process is iterative, aiming to bring the simplex closer to the minimum of the objective function which is calculated in equation 2. The algorithm assesses the objective function at specific points in the simplex at each iteration, and then updates the simplex by applying a number of transformations, such as reflection, contraction, expansion, and shrinkage. These changes help gradually improve the simplex towards the best solution. They systematically explore the parameter space, which is determined by the manufacturer's specified minimum and maximum values for each bias. The algorithm continues iterating until convergence criteria are met, such as reaching a specified number of iterations or achieving a desired level of accuracy in the objective function value \cite{m10}. As previously stated, the criteria in this study is the required modification of each bias. The target function which is defined as follows:

\begin{equation}
 Target Function = 1- Efficacy Metric
\end{equation}

and it is equivalent as maximizing the efficacy metric that was determined in the previous step. As a result, the optimization process increases the amount of frames that have a face detected on them.

\begin{figure}[ht]
  \centering
  \includegraphics[width=9cm]{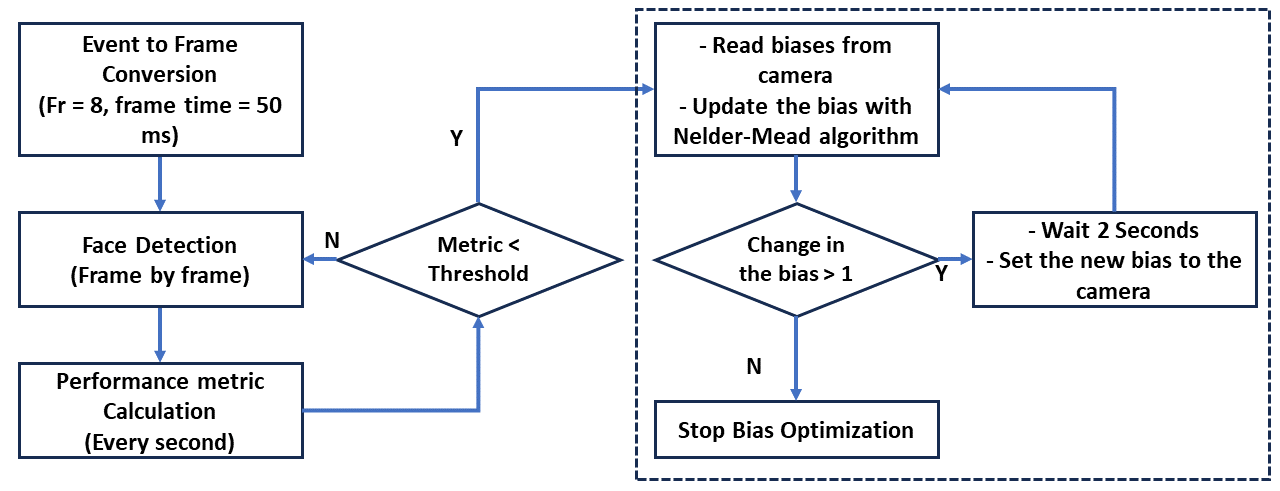}
  \caption{Autobiasing system flowchart; 'Bias Controller' enclosed within the dashed box.}
  \label{fig:3}
\end{figure}

The flowchart of the autobiasing pipeline is illustrated in Figure \ref{fig:3}.

\section{Results and Discussion}
In order to demonstrate the effectiveness of the proposed technique and verify its capacity to optimize multiple bias values, it was tested in presence of two major factors that negatively impact lighting conditions: flickering and darkness. According to the test results, which will be discussed, face detection with the event camera's default biases is impossible due to these factors. During the experiments the illumination level remained below 200 lux, as measured by the SEKONIC C-500 light meter in an office environment with only natural illumination, with no artificial light sources. Flickering was produced using four infrared LEDs. The flashing frequencies of these LEDs were controlled by an AC signal generator and ranged from 100 Hz to 500 Hz. Three human participants gave their informed consent and were positioned for 180 seconds at each flickering frequency, 60 cm distant from the Prophesee IMX 636 event camera. The tests aimed to determine if the autobiasing system could adjust the camera bias settings to enable the face detection system to identify faces. During the first 20 seconds of the test, the autobiasing system was turned off, and the event camera operated with its default bias values which are zero. Under these lighting conditions and bias settings, face detection was nearly impossible.

\begin{figure}[ht!]
    \centering
    \begin{subfigure}[b]{0.3\textwidth}
        \includegraphics[width=\textwidth]{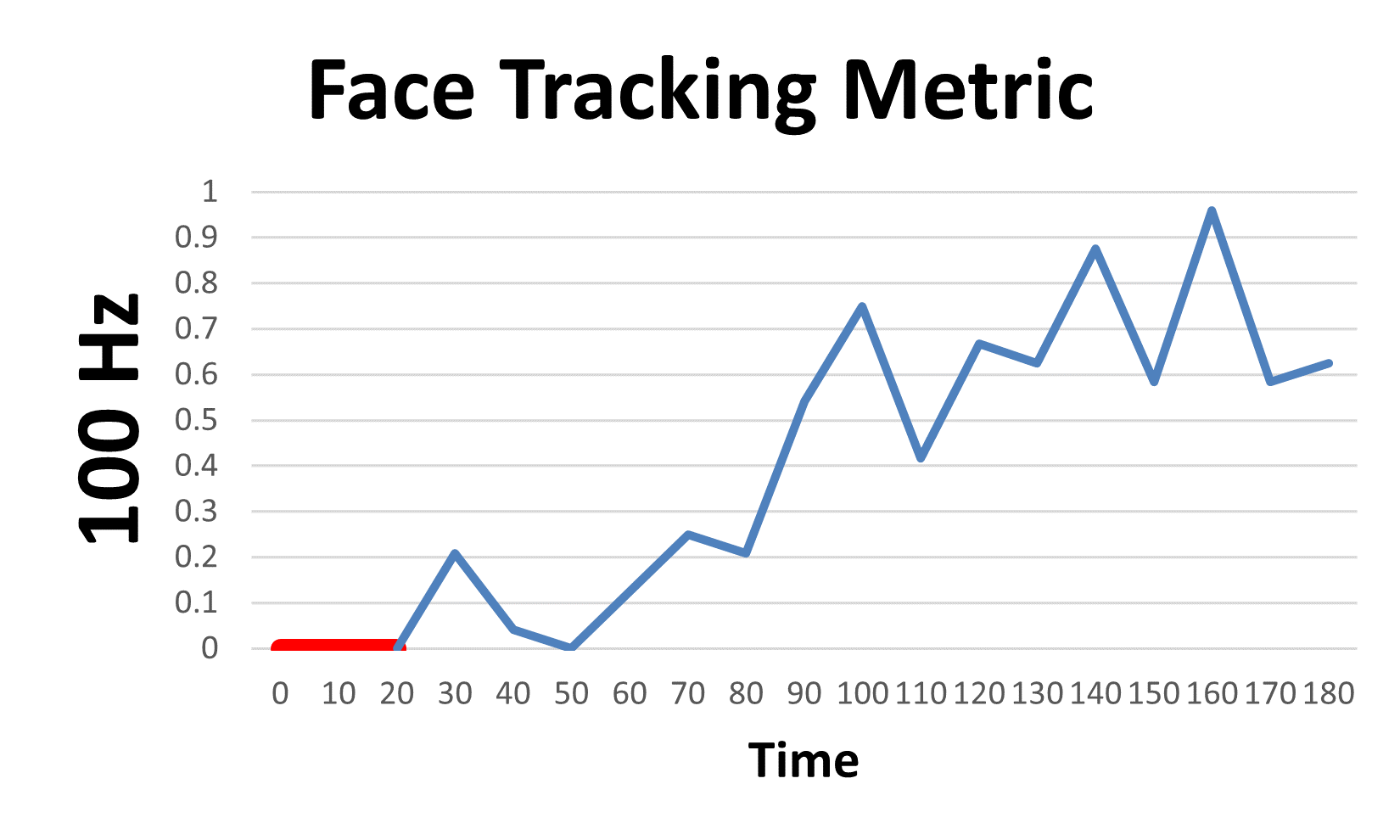}
    \end{subfigure}%
    \begin{subfigure}[b]{0.3\textwidth}
        \includegraphics[width=\textwidth]{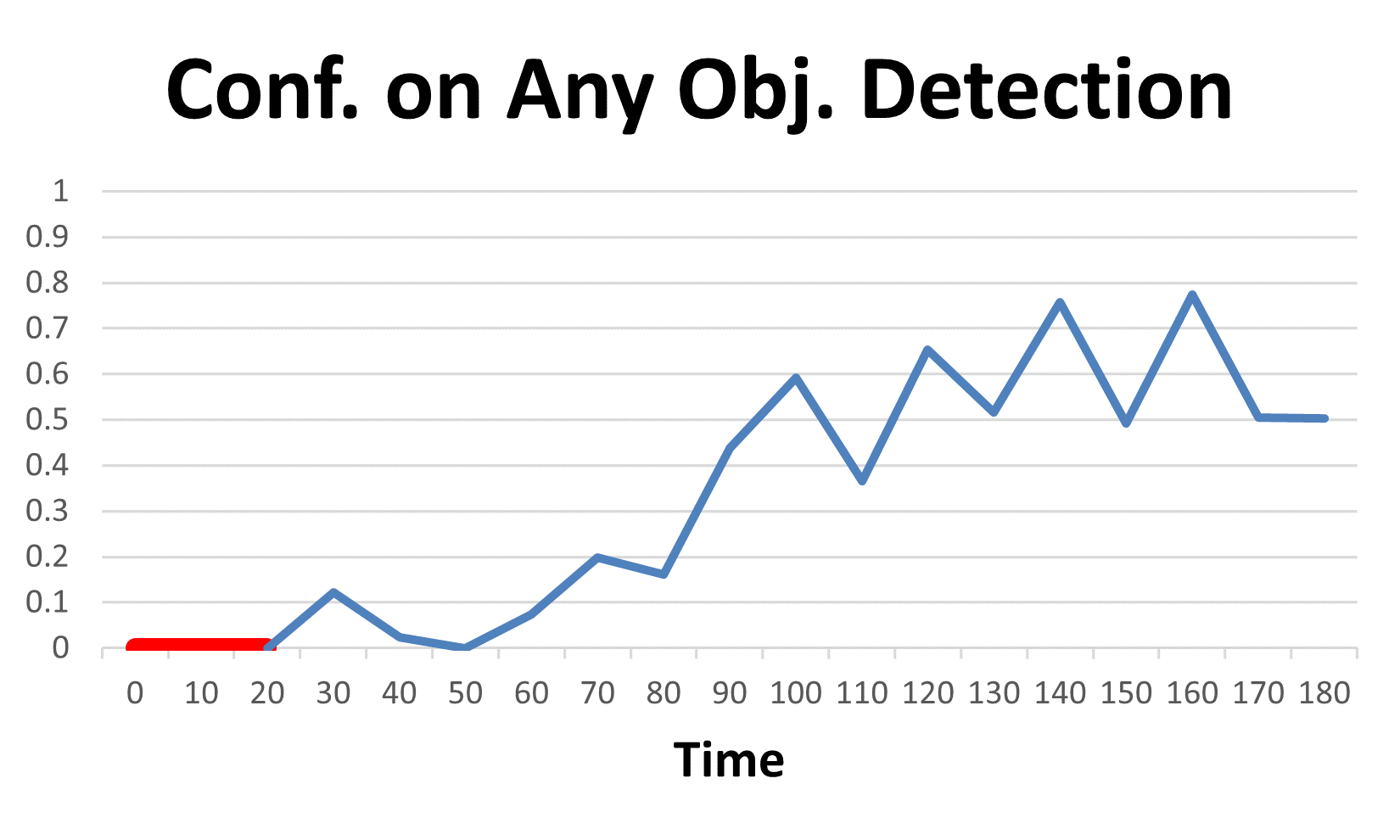}
    \end{subfigure}%
    \begin{subfigure}[b]{0.3\textwidth}
        \includegraphics[width=\textwidth]{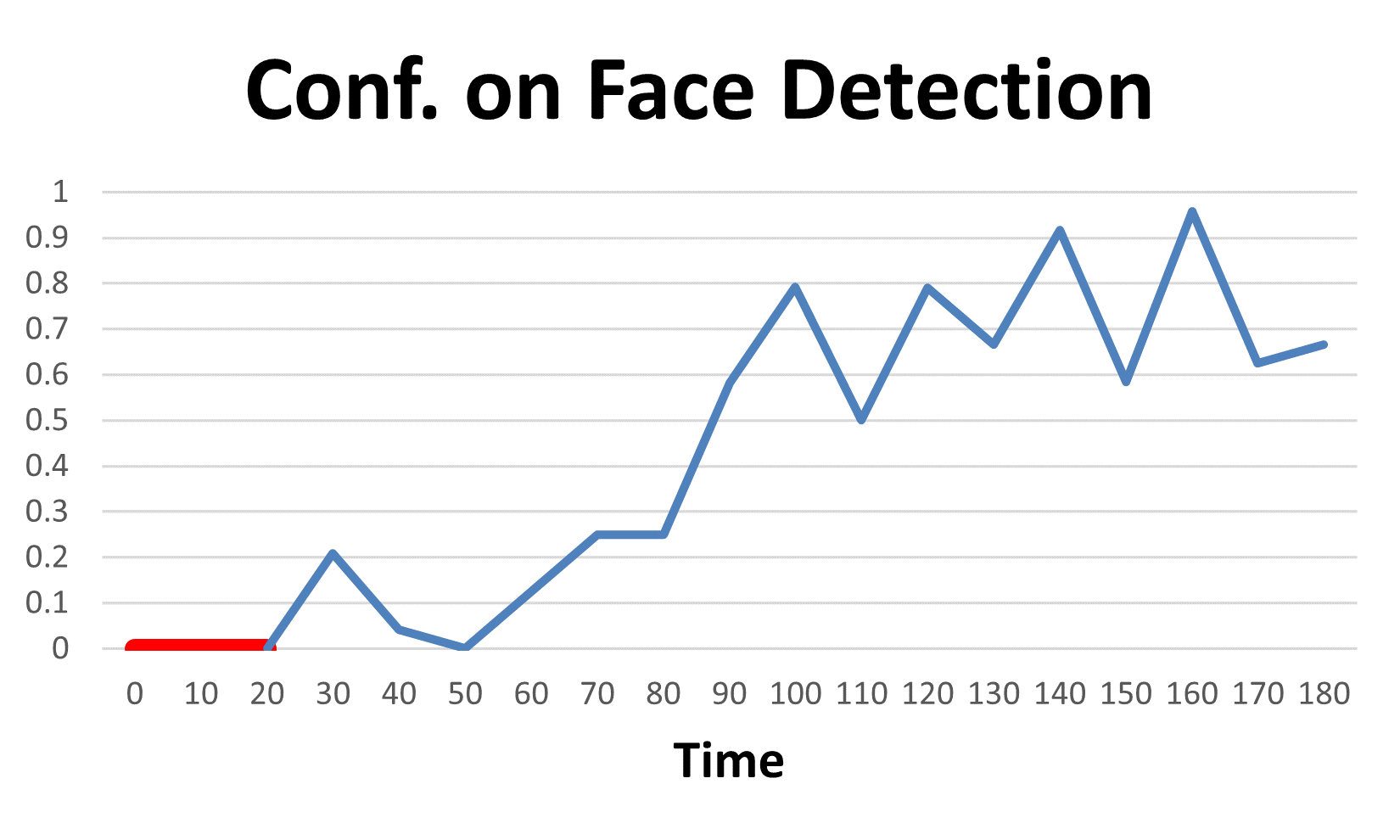}
    \end{subfigure}%

    \begin{subfigure}[b]{0.3\textwidth}
        \includegraphics[width=\textwidth]{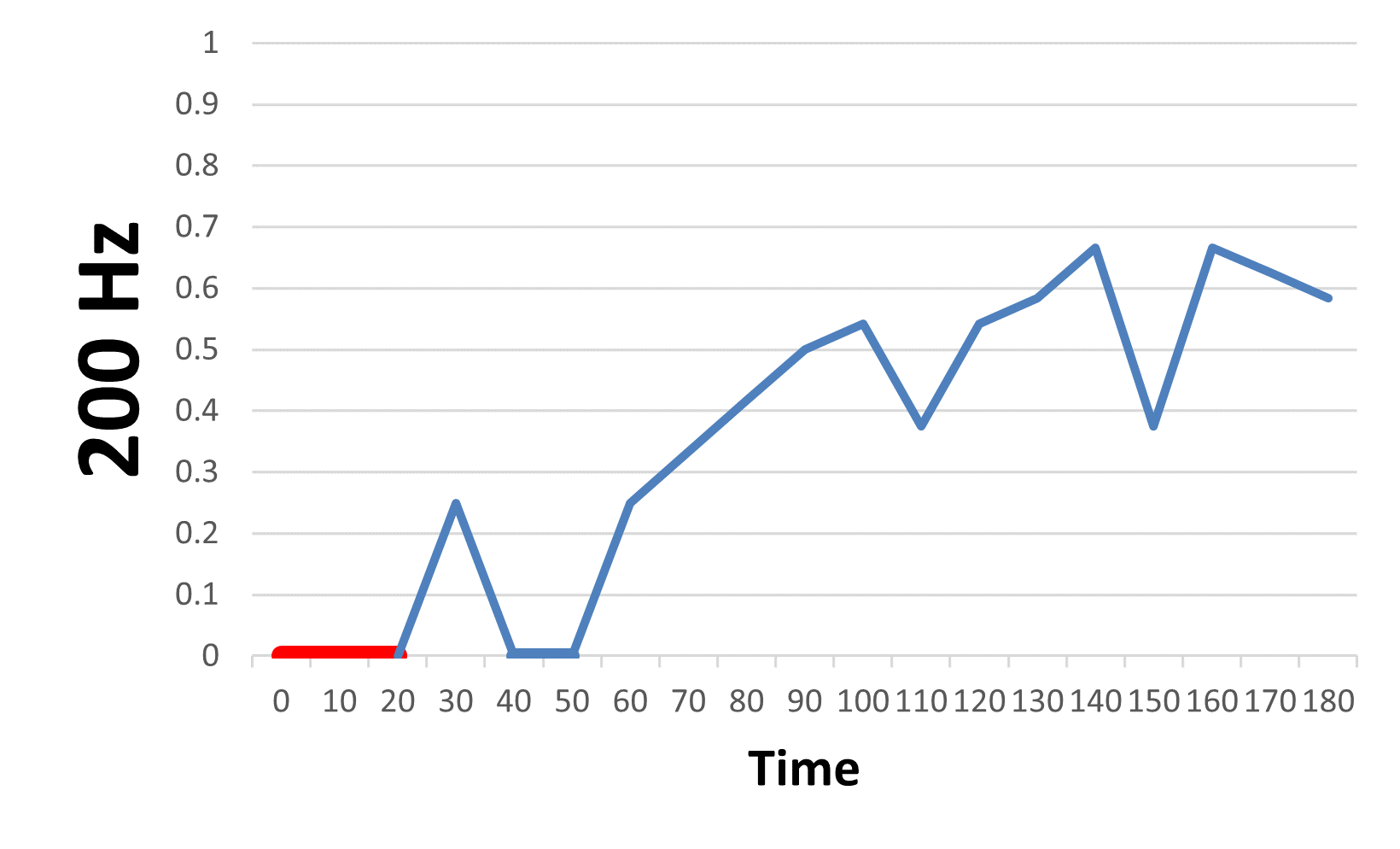}
    \end{subfigure}%
    \begin{subfigure}[b]{0.3\textwidth}
        \includegraphics[width=\textwidth]{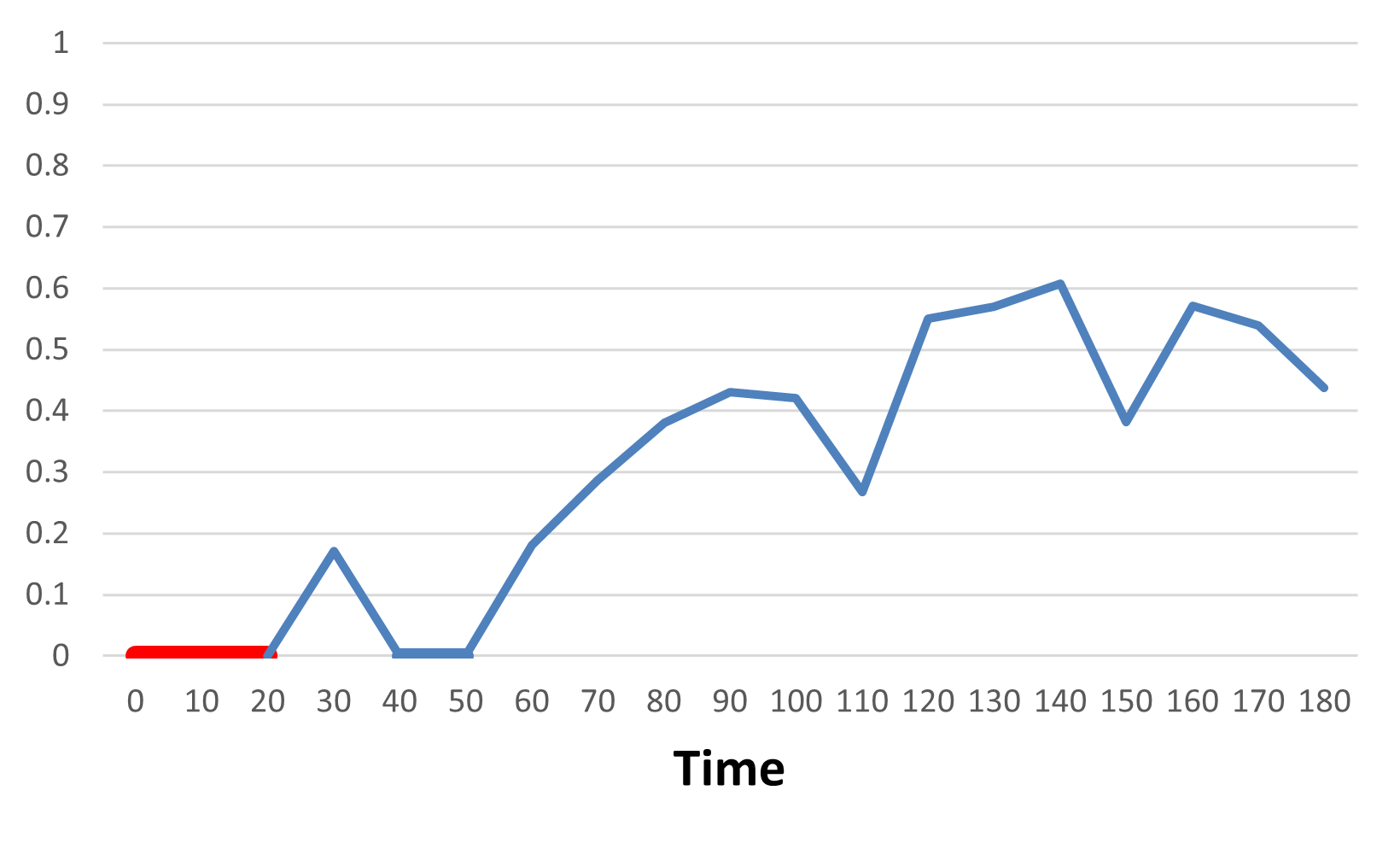}
    \end{subfigure}%
    \begin{subfigure}[b]{0.3\textwidth}
        \includegraphics[width=\textwidth]{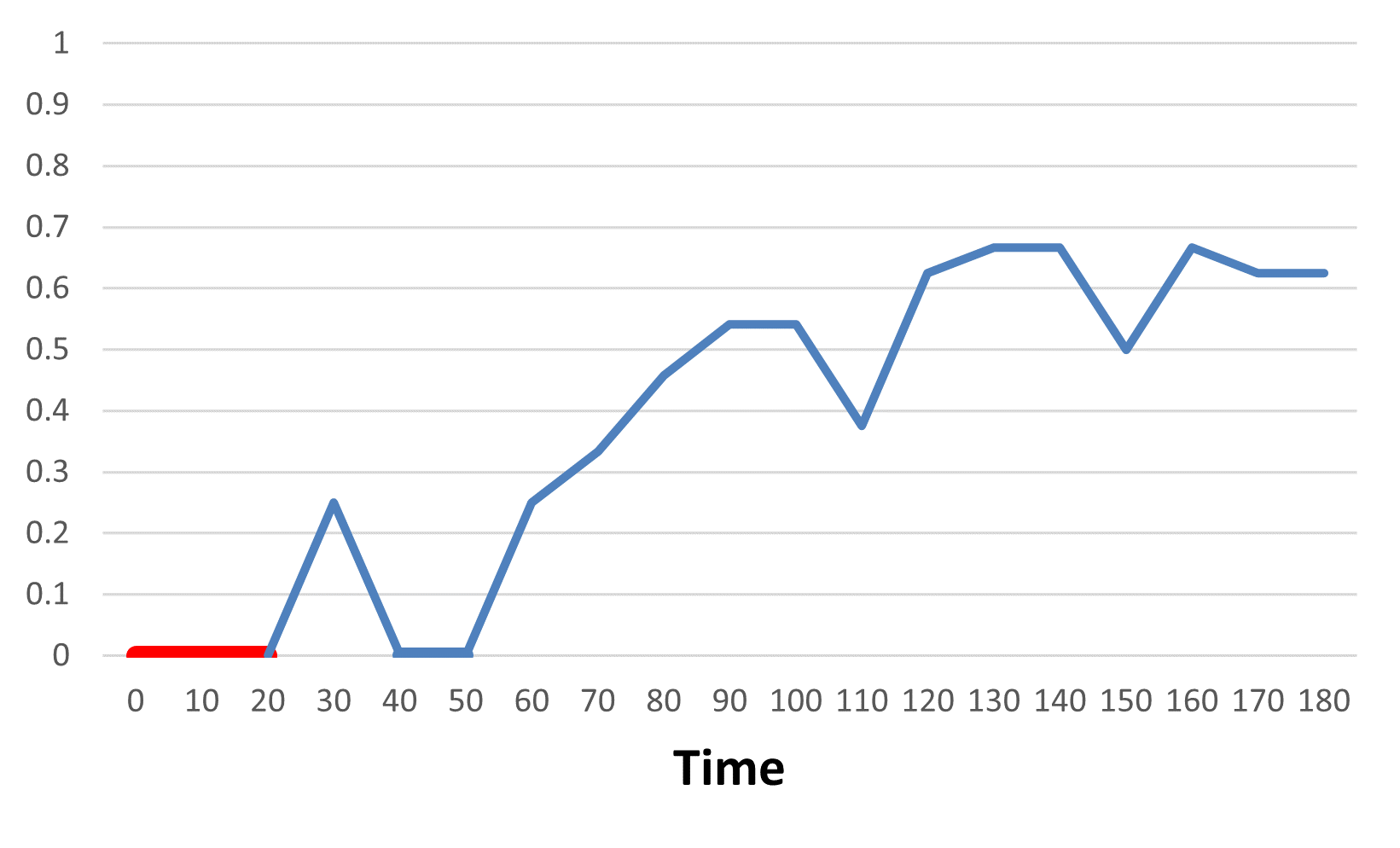}
    \end{subfigure}%

    \begin{subfigure}[b]{0.3\textwidth}
        \includegraphics[width=\textwidth]{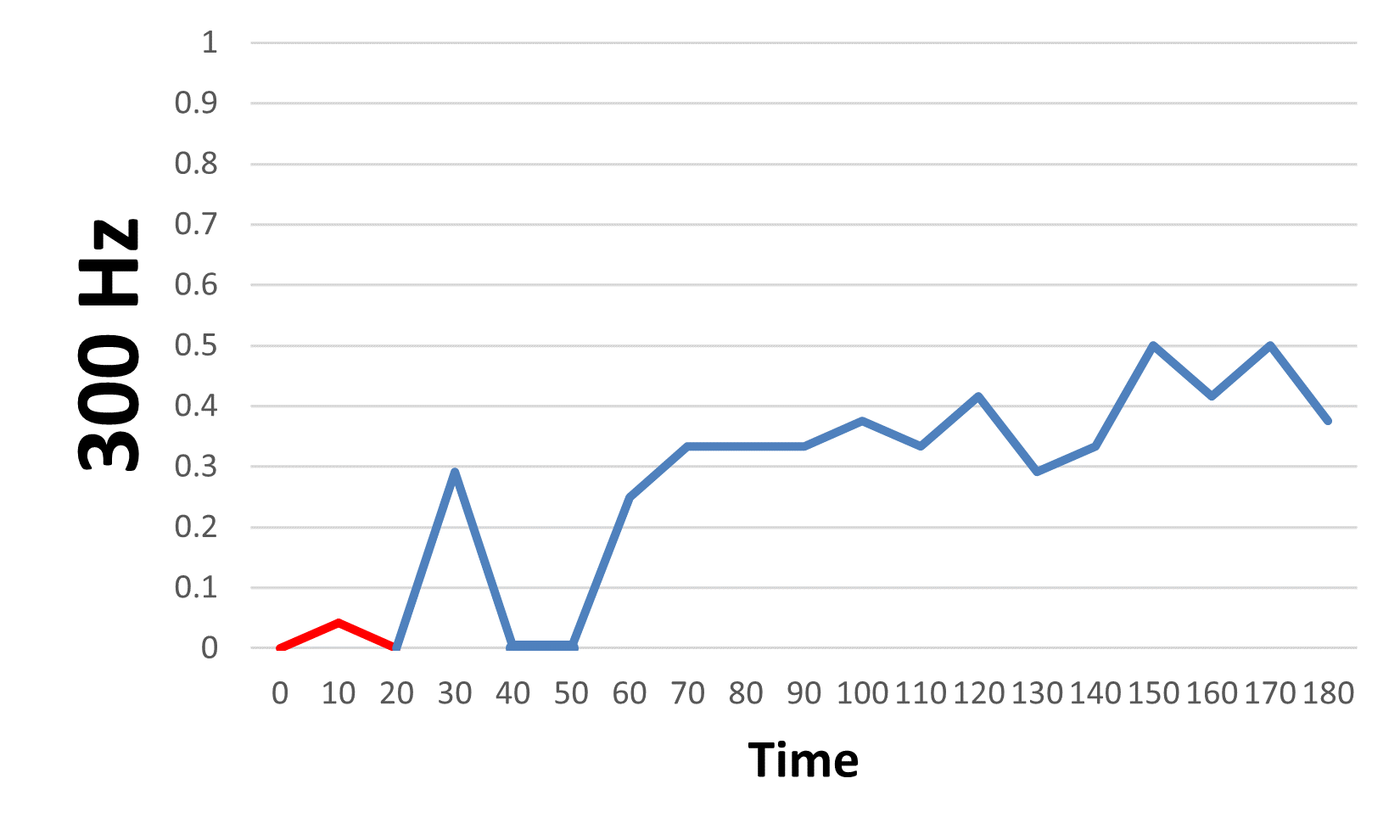}
    \end{subfigure}%
    \begin{subfigure}[b]{0.3\textwidth}
        \includegraphics[width=\textwidth]{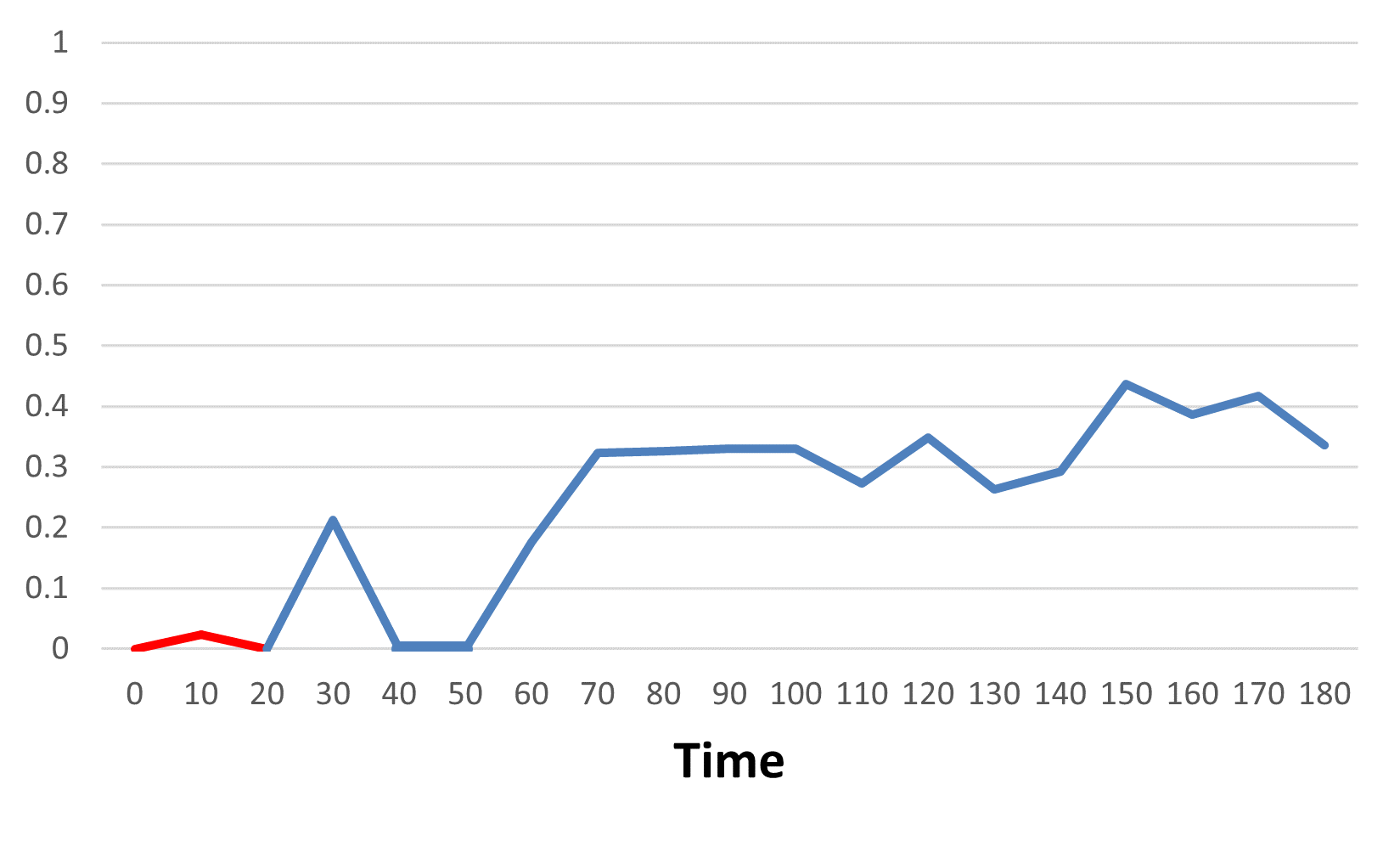}
    \end{subfigure}%
    \begin{subfigure}[b]{0.3\textwidth}
        \includegraphics[width=\textwidth]{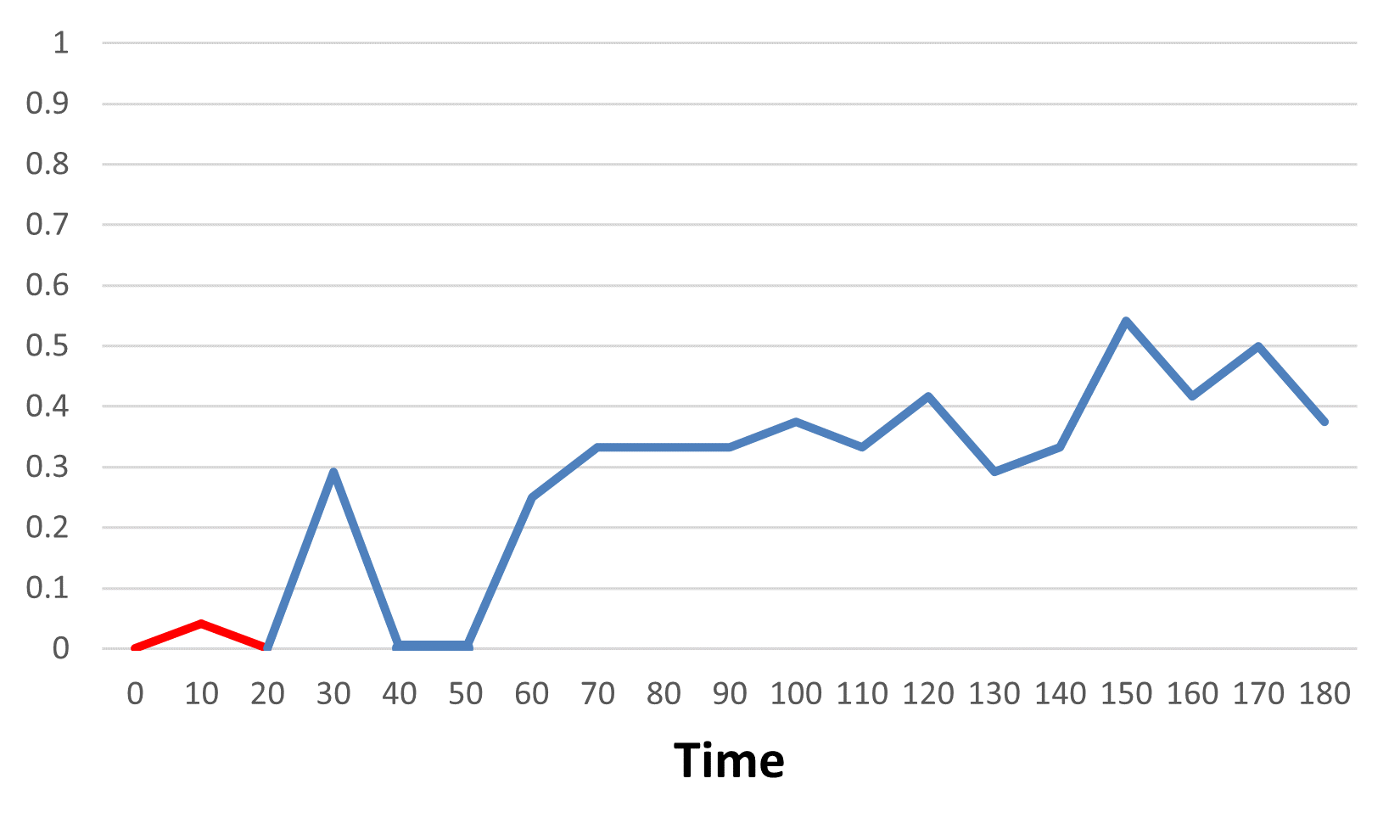}
    \end{subfigure}%

    \begin{subfigure}[b]{0.3\textwidth}
        \includegraphics[width=\textwidth]{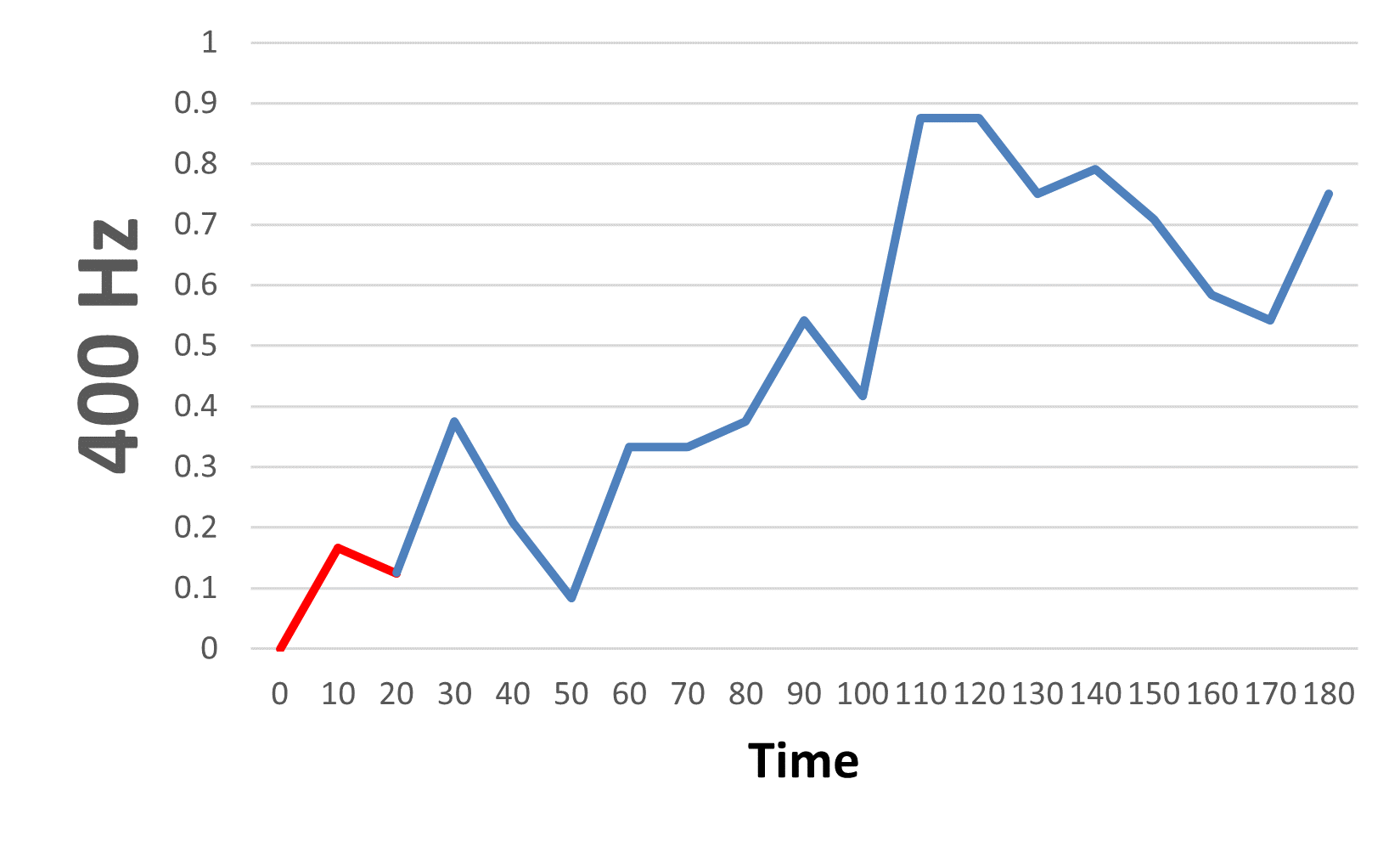}
    \end{subfigure}%
    \begin{subfigure}[b]{0.3\textwidth}
        \includegraphics[width=\textwidth]{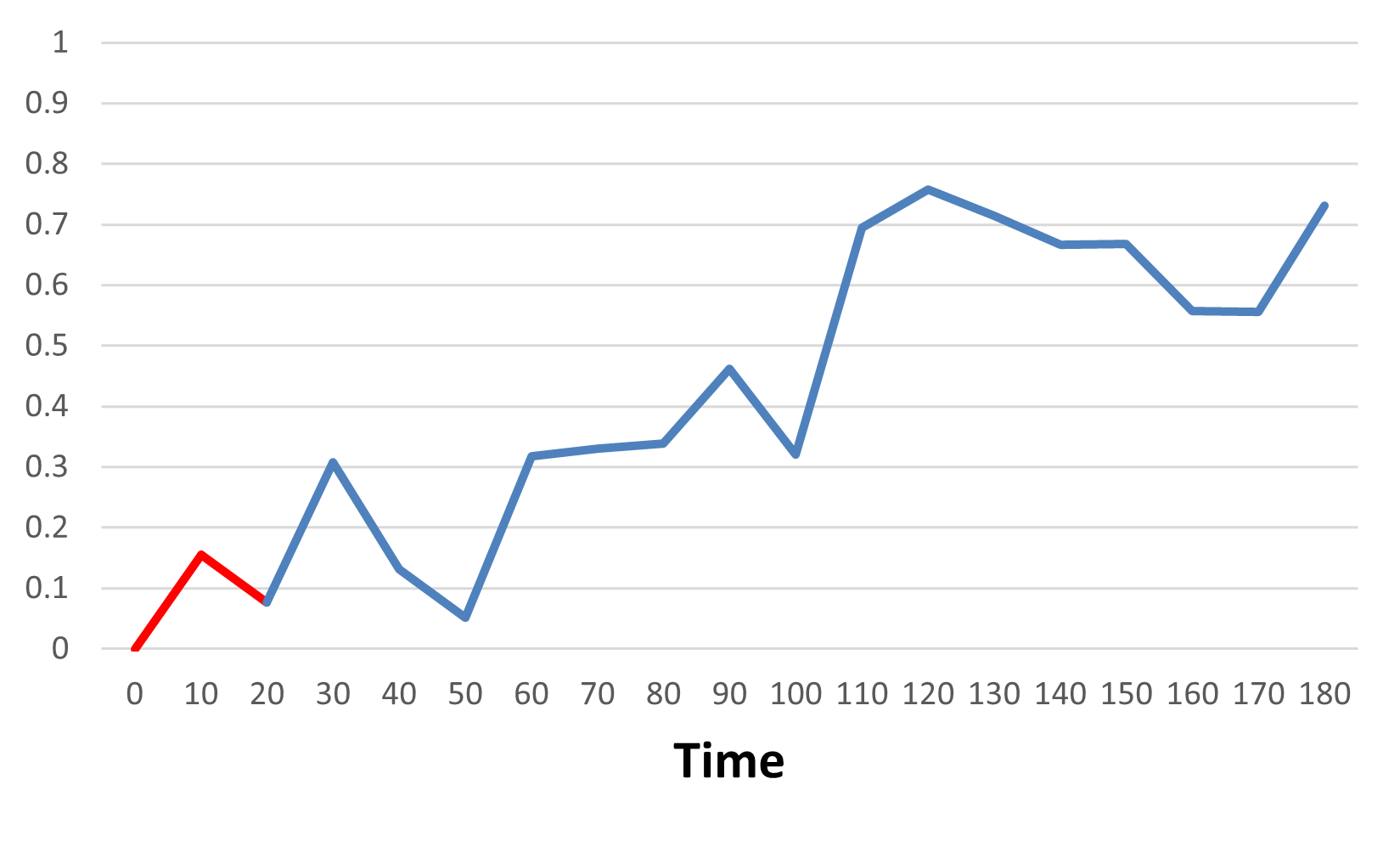}
    \end{subfigure}%
    \begin{subfigure}[b]{0.3\textwidth}
        \includegraphics[width=\textwidth]{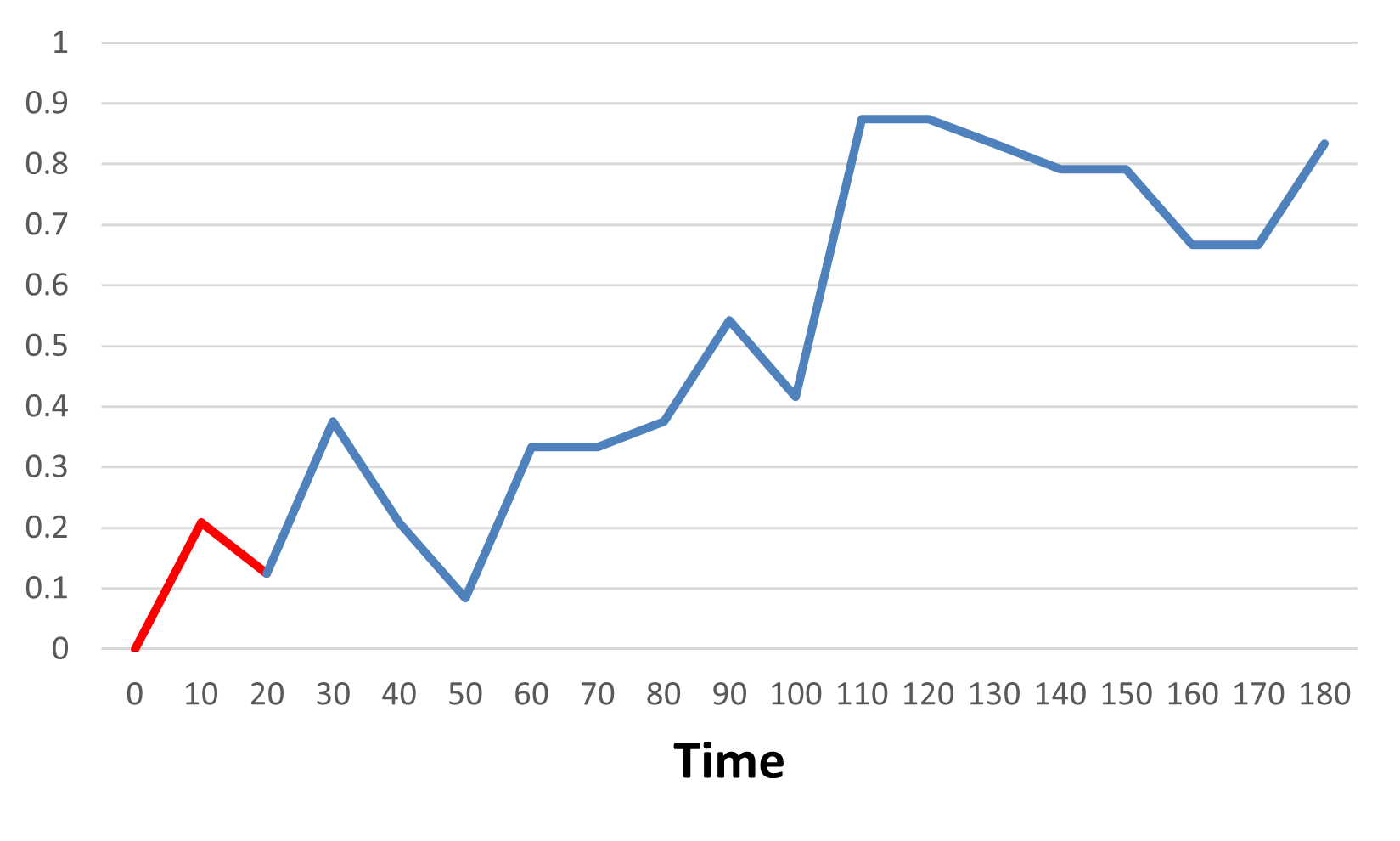}
    \end{subfigure}%

    \begin{subfigure}[b]{0.3\textwidth}
        \includegraphics[width=\textwidth]{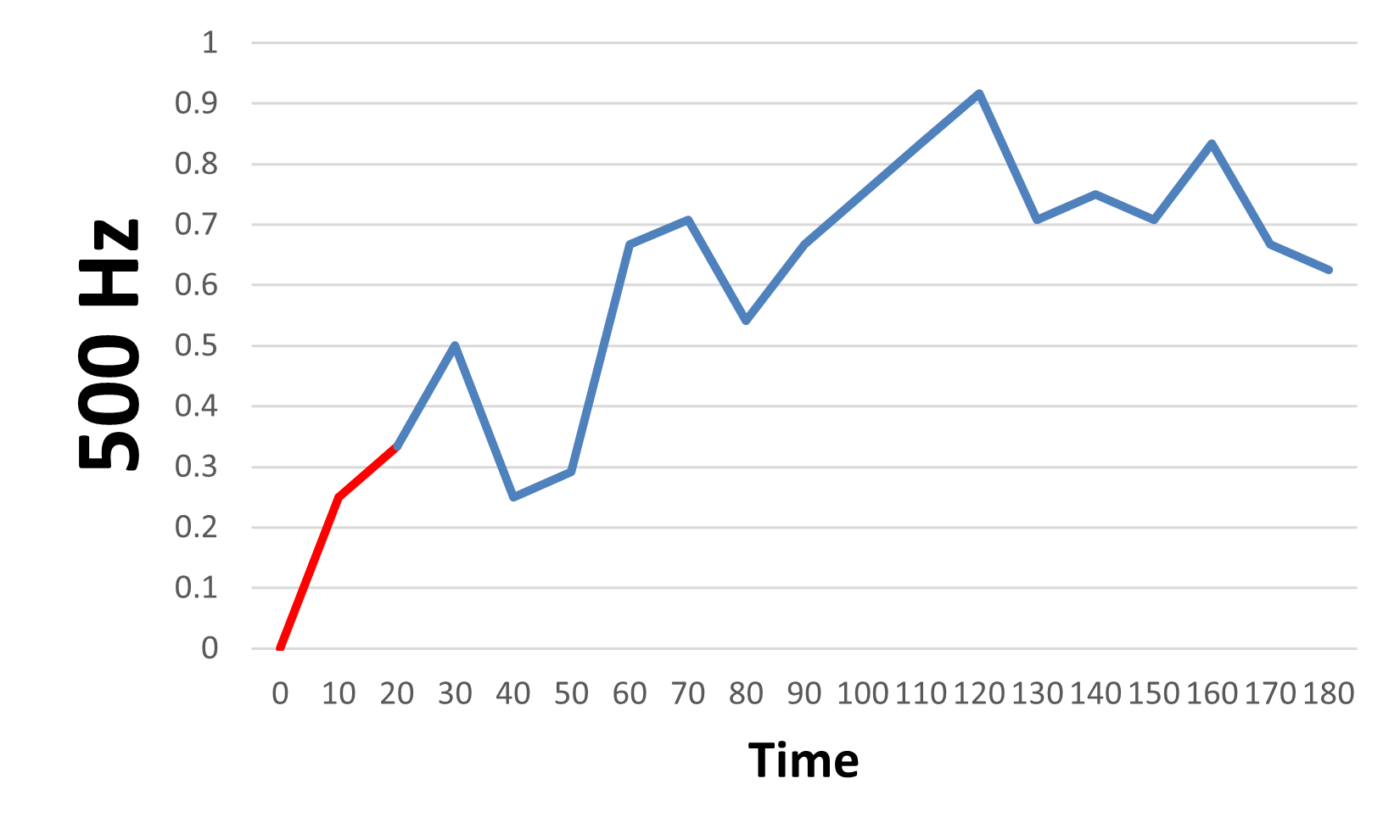}
    \end{subfigure}%
    \begin{subfigure}[b]{0.3\textwidth}
        \includegraphics[width=\textwidth]{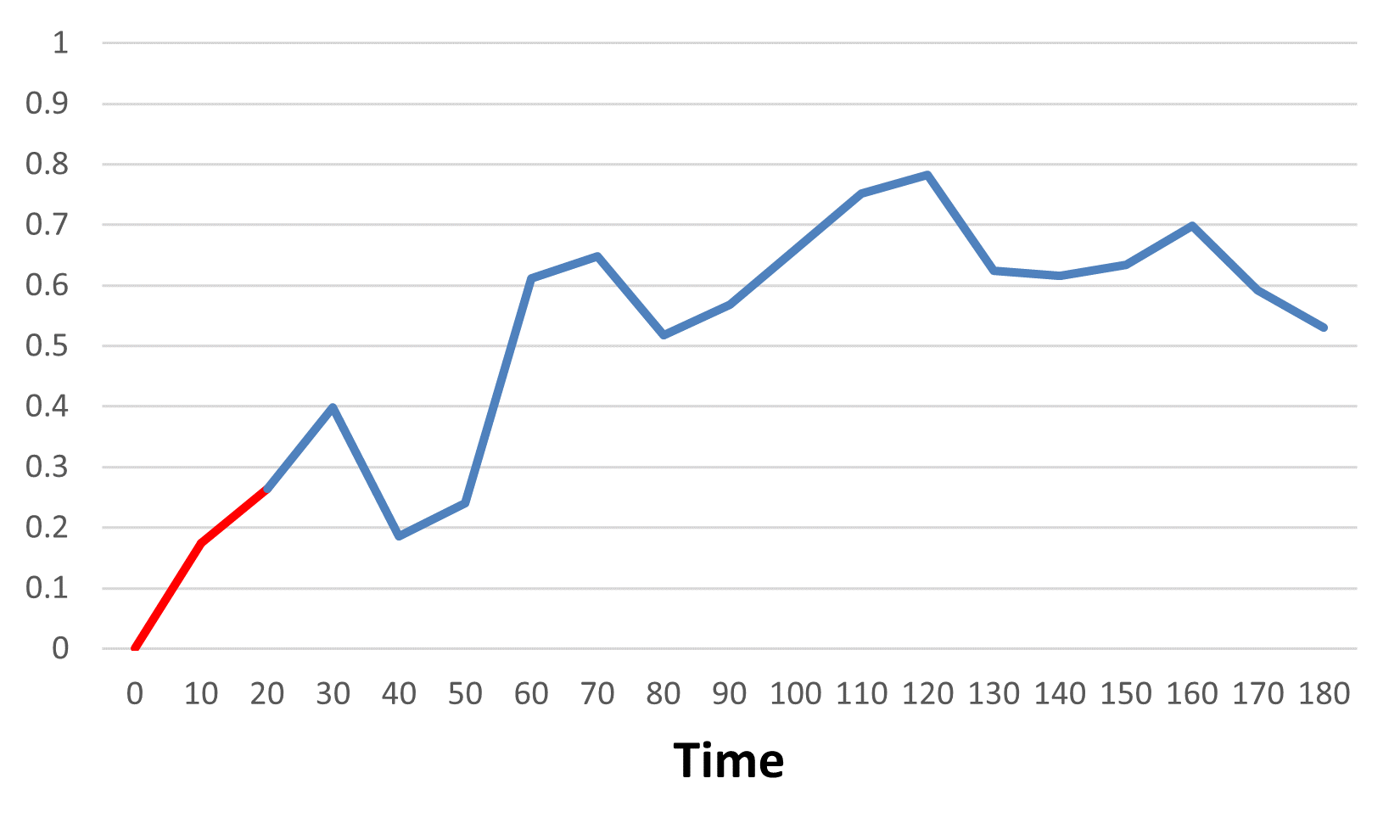}
    \end{subfigure}%
    \begin{subfigure}[b]{0.3\textwidth}
        \includegraphics[width=\textwidth]{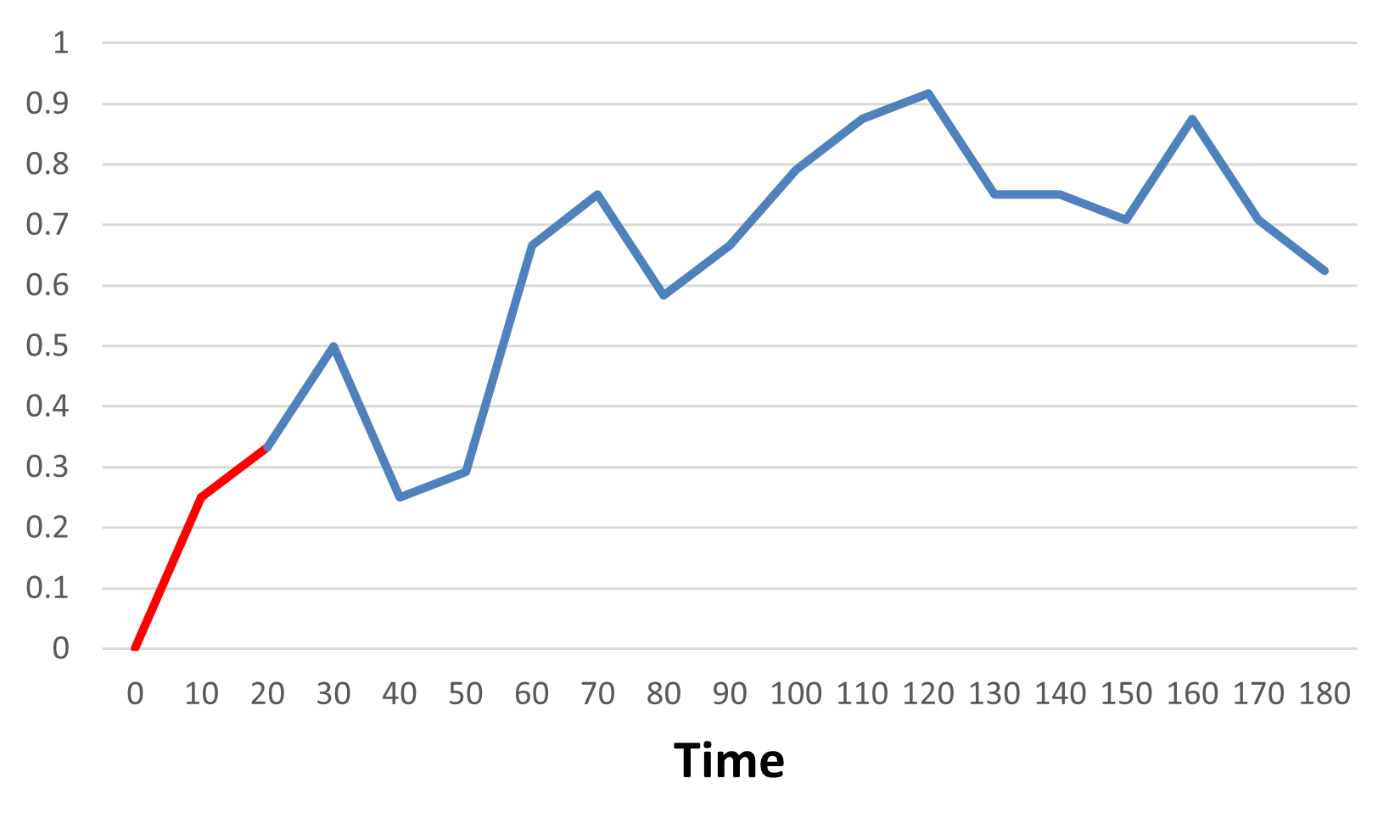}
    \end{subfigure}%

    \caption{Impact of autobiasing on face tracking metrics under 200 lux and flickering frequencies from 100 Hz to 500 Hz (Rows: flickering frequencies, Columns: face tracking metric, YOLO confidences on both object and face detection.)}
    \label{fig:4}
\end{figure}

Three metrics are utilized to provide a quantitative basis to evaluate the autobiasing system's performance. These include YOLO confidence in both face and any object detection, as well as the face tracking efficacy metric described in previous sections. The average amount of the metrics during the autobiasing process for all representative participants is depicted in Figure \ref{fig:4}. The first 20 seconds of all graphs are colored in red and depicts the period at which the camera works in the default bias setting and the autobiasing has not begun yet. As can be seen in all graphs the face detection is either impossible or very weak in all conditions at this red area. However, the detection confidence increases once the autobiasing begins, continuing until the biases reach their optimum values. The increase in all metric values across the graphs demonstrates that autobiasing significantly improves the performance of an event-based vision application under adverse lighting conditions. An important point to consider is that the time required to reach the optimal bias settings depends on the chosen frame rate and can be reduced if application requirements or computational resources permit.

To better understand the impact of autobiasing on face tracking metrics, Table \ref{tbl:improvements} shows the average percentage change in each metric, calculated across all subjects for each frequency. As can be seen in the table, all metrics have increased after autobiasing.

\begin{table}[t]
    \centering
    \fontsize{7pt}{9pt}\selectfont
    \caption{Percentage of alterations in diverse metrics subsequent to autobiasing under low-light conditions and various flickering frequencies (Lux < 200).}
        \begin{tabular}{|c|c|c|c|}
            \hline
            \multirow{2}{*}{\textbf{Flickering Frequency (Hz)}} & \multicolumn{2}{|c|}{\textbf{YOLO Confidence (\%)}} & \multirow{2}{*}{\textbf{Face tracking metric(\%)}} \\
            \cline{2-3}
            & Any Obj. & Face & \\
            \hline
            100 & +50 & +66 & +62 \\
            \hline
            200 & +43 & +62 & +58 \\
            \hline
            300 & +33 & +37 & +37 \\
            \hline
            400 & +73 & +83 & +75 \\
            \hline
            500 & +53 & +62 & +62 \\
            \hline
        \end{tabular}
        \label{tbl:improvements}

\end{table}

The average bias values after 180 seconds for each frequency are presented in Table \ref{tbl:1}. The maximum changes occurred for bias\_diff\_off and bias\_fo. These biases were expected to change the most under flickering and darkness as the bias\_diff\_off sets the negative light threshold and was expected to increase in a dark environment, while bias\_fo mitigates flickering and background noise, with lower values reducing flickering. This indicates that the algorithm correctly identified the biases needing adjustment and made appropriate changes.

\newcolumntype{P}[1]{>{\centering\arraybackslash}p{#1}}

\begin{table}[h]
    \centering
    \fontsize{7pt}{9pt}\selectfont
    \caption{Bias values after 180 seconds in different flickering frequencies and lighting under 200 lux}
    \begin{tabular}{|P{2.4 cm}|P{1 cm}|P{1.1 cm}|P{1.1 cm}|P{1.1 cm}|P{1.1 cm}|}
        \hline
        \multirow{2}{*}{\textbf{Frequency (Hz)}} & \multicolumn{5}{|c|}{\textbf{Average Bias Values After Autobiasing}}  \\
        \cline{2-6}
        & diff\_off & diff\_on & fo & hpf & refr  \\
        \hline
        100 & 35 & -2 & -3 & 0 & -2 \\
        \hline
        200 & 29 & 0 & -10 & 2 & 8 \\
        \hline
        300 & 26 & 0 & -3 & 1 & 1 \\
        \hline
        400 & 33 & 11 & -22 & 11 & 0 \\
        \hline
        500 & 25 & 5 & -14 & 7 & -3 \\
        \hline
    \end{tabular}
    \label{tbl:1} 
\end{table}

The proposed method is the first algorithm that tunes all biases simultaneously. Previous research even on single bias tuning is limited. Prior studies have focused on optimizing the number of generated events to address memory and computational concerns and were developed for different applications than the case study of this paper. As a result, making a numerical comparison between the proposed method and others is not possible. However, Table \ref{tbl:2} offers a descriptive comparison of the automatic bias systems.


\begin{table*}
\centering
\fontsize{7pt}{9pt}\selectfont
\caption{Comparison of various autobiasing methods}
\begin{tabular}{|C{2.4cm}|C{3cm}|C{2cm}|C{4cm}|}
\hline
\textbf{Method} & \textbf{Optimized Biases} & \textbf{Mode} & \textbf{Application} \\
\hline
\cite{m11} & Single: bias\_ref & Real-time & Visual place recognition \\
\hline
\cite{w19} & Single: bias\_ref & Real-time & Not application specific \\
\hline
Proposed method & All biases & Real-time & Not application specific  \\
\hline
\end{tabular}
\label{tbl:2}
\end{table*}

\subsection{Limitations}
Despite its robust performance, the proposed autobiasing system could be improved by employing AI to better adapt to new lighting conditions. Exploring the trade-off between achieving faster and more accurate results at the expense of increased complexity is worthwhile. Furthermore, the testing was designed to be as challenging as possible to demonstrate the system's capabilities. However, numerous lighting conditions of automotive applications are not possible to study within the limitations of a single research paper.

\section{Conclusion}
Event cameras offer significant advantages over frame-based sensors by capturing events with higher temporal resolution, as they independently record changes in light at each pixel. However, their performance can be inconsistent under varying lighting conditions, posing challenges for tasks such as face detection in dynamic environments such as in-cabin monitoring. In this paper, the problem is addressed by utilizing the event cameras' built-in adjustable bias settings. The proposed approach involves continuous monitoring of the event-based DMS's performance. If face detection, a critical component of the DMS, begins to malfunction, the camera's bias settings are dynamically adjusted until the application functions satisfactorily. Monitoring of face detection performance is done by utilizing YOLO confidences for both face and object detection, and an additional metric that indicates the success of the face detector over time. To optimize the bias values for face detection performance modification, a simplex-based technique known as Nelder-Mead is employed. The system was tested under challenging conditions, including low light levels (less than 200 lux) and flickering frequencies (100 Hz to 500 Hz). Under these conditions, face detection and tracking were completely impossible with the default bias settings. Post-autobiasing, there were significant improvements: YOLO confidences increased by at least 33\% for object detection, 37\% for face detection, and 37\% for face tracking.

This paper has proposed a novel autobiasing method for dynamically improving the performance of a class of time event algorithms in real time. It is the first approach which optimizes all biases simultaneously for this purpose, thus enhancing performance in multiple lighting conditions. Future research will explore advanced optimization techniques, including AI-based methods such as reinforcement learning, to further enhance adaptability and efficiency without repetitive optimization.



%
%
\bibliographystyle{splncs04}
\bibliography{main}

\begin{thebibliography}{10}
\providecommand{\url}[1]{\texttt{#1}}
\providecommand{\urlprefix}{URL }
\providecommand{\doi}[1]{https://doi.org/#1}

\bibitem{m17}
Bergstra, J., Bengio, Y.: Random search for hyper-parameter optimization. Journal of Machine Learning Research  \textbf{13},  281--305 (2012)

\bibitem{w19}
Delbruck, T., Graca, R., Paluch, M.: Feedback control of event cameras. In: 2021 IEEE/CVF Conference on Computer Vision and Pattern Recognition Workshops (CVPRW). pp. 1324--1332. IEEE (2021). \doi{10.1109/CVPRW53098.2021.00146}

\bibitem{m5}
Dilmaghani, M.S., Shariff, W., Ryan, C., Lemley, J., Corcoran, P.: {Control and evaluation of event cameras output sharpness via bias}. In: Osten, W., Nikolaev, D.P., Zhou, J.J. (eds.) Fifteenth International Conference on Machine Vision (ICMV 2022). vol. 12701, p. 127011I. International Society for Optics and Photonics, SPIE (2023). \doi{10.1117/12.2679755}, \url{https://doi.org/10.1117/12.2679755}

\bibitem{m1}
Gallego, G., Delbrück, T., Orchard, G., Bartolozzi, C., Taba, B., Censi, A., Leutenegger, S., Davison, A.J., Conradt, J., Daniilidis, K., Scaramuzza, D.: Event-based vision: A survey. IEEE Transactions on Pattern Analysis and Machine Intelligence  \textbf{44}(1),  154--180 (2022). \doi{10.1109/TPAMI.2020.3008413}

\bibitem{w5}
Gao, N., Zhang, Z., Deng, J., Guo, X., Cheng, B., Hou, H.: Acoustic metamaterials for noise reduction: A review. Advanced Materials Technologies  \textbf{7} (2022). \doi{10.1002/admt.202100698}

\bibitem{w1}
Gomez, A.L., Saravi, S., Edirisinghe, E.: Multiexposure and multifocus image fusion with multidimensional camera shake compensation. Optical Engineering  \textbf{52} (2013). \doi{10.1117/1.OE.52.10.102007}

\bibitem{w17}
Graca, R., Delbruck, T.: Unraveling the paradox of intensity-dependent dvs pixel noise. In: 2021 International Image Sensor Workshop (IISW). IEEE (2021). \doi{10.48550/ARXIV.2109.08640}

\bibitem{w18}
Graca, R., McReynolds, B., Delbruck, T.: Optimal biasing and physical limits of dvs event noise. In: 2023 International Image Sensor Workshop (IISW). IEEE (2023). \doi{10.48550/arXiv.2304.04019}

\bibitem{m13}
Hansen, N., Ostermeier, A.: Completely derandomized self-adaptation in evolution strategies. Evolutionary Computation  \textbf{9}(2),  159--195 (2001)

\bibitem{m12}
Holland, J.H.: Adaptation in Natural and Artificial Systems. University of Michigan Press (1975)

\bibitem{w23}
Hu, Y., Liu, S.C., Delbruck, T.: V2e: From video frames to realistic dvs events. In: Proceedings of the IEEE/CVF Conference on Computer Vision and Pattern Recognition (CVPR) Workshops. pp. 1312--1321 (Jun 2021)

\bibitem{w3}
Jia-wen, L.: Survey of the auto-focus methods based on image processing. Laser \& Infrared  (2013)

\bibitem{w24}
Joubert, D., Marcireau, A., Ralph, N., Jolley, A., van Schaik, A., Cohen, G.: Event camera simulator improvements via characterized parameters. Frontiers in Neuroscience  \textbf{15} (2021). \doi{10.3389/fnins.2021.702765}

\bibitem{w6}
Kaur, N., Singh, E.K.: Image enhancement techniques: A selected review. International journal of engineering research and technology  \textbf{2} (2013)

\bibitem{m14}
Kennedy, J., Eberhart, R.: Particle swarm optimization. In: Proceedings of IEEE International Conference on Neural Networks. vol.~4, pp. 1942--1948. IEEE (1995)

\bibitem{m15}
Kirkpatrick, S., Gelatt~Jr, C.D., Vecchi, M.P.: Optimization by simulated annealing. Science  \textbf{220}(4598),  671--680 (1983)

\bibitem{w10}
Koli, M., Balaji, S.: Literature survey on impulse noise reduction. Signal \& Image Processing : An International Journal  \textbf{4},  75--95 (2013). \doi{10.5121/SIPIJ.2013.4506}

\bibitem{w9}
Konnik, M., Welsh, J.: High-level numerical simulations of noise in ccd and cmos photosensors: review and tutorial. ArXiv  \textbf{abs/1412.4031} (2014)

\bibitem{w16}
Li, C.: Two-stream vision sensors. Ph.D. thesis, Ph.D. dissertation (2017)

\bibitem{w14}
Lichtsteiner, P.: An AER temporal contrast vision sensor. Ph.D. thesis, Ph.D. dissertation (2006)

\bibitem{w11}
Lichtsteiner, P., Posch, C., Delbruck, T.: A 128x128 120 db 15 µs latency asynchronous temporal contrast vision sensor. IEEE Journal of Solid-State Circuits  \textbf{43}(2),  566--576 (2008)

\bibitem{w4}
Mallick, T., Das, P., Majumdar, A.: Characterizations of noise in kinect depth images: A review. IEEE Sensors Journal  \textbf{14},  1731--1740 (2014). \doi{10.1109/JSEN.2014.2309987}

\bibitem{w21}
McReynolds, B.J., Graca, R.P., Delbruck, T.: Experimental methods to predict dynamic vision sensor event camera performance. Optical Engineering  \textbf{61}(7),  074103 (2022). \doi{10.1117/1.OE.61.7.074103}

\bibitem{m4}
Metavision sdk docs 4.5.2 documentation, \url{https://docs.prophesee.ai/stable/hw/manuals/biases.html}

\bibitem{w13}
Biases — metavision intelligence docs 3.1.2 documentation. \url{https://docs.prophesee.ai/stable/hw/manuals/biases.html} (2023), accessed: 2023-3-17

\bibitem{w15}
Moeys, D.P.: Analog and digital implementations of retinal processing for robot navigation systems. Ph.D. thesis, Ph.D. dissertation (2016)

\bibitem{w7}
Mudhafar, R.A.A., Abbadi, N.K.E.: Noise in digital image processing: A review study. 2022 3rd Information Technology To Enhance e-learning and Other Application (IT-ELA) pp. 79--84 (2022). \doi{10.1109/IT-ELA57378.2022.10107965}

\bibitem{m11}
Nair, G.B., Milford, M., Fischer, T.: Enhancing visual place recognition via fast and slow adaptive biasing in event cameras (2024)

\bibitem{w8}
Prasad, P., Anitha, D., Anil, D.: A systematic review of noise types, denoising methods, and evaluation metrics in images. 2023 IEEE International Conference on Recent Advances in Systems Science and Engineering (RASSE) pp.~1--9 (2023). \doi{10.1109/RASSE60029.2023.10363591}

\bibitem{m8}
 (Mar 2024), \url{https://www.prophesee.ai/event-camera-evk4/}

\bibitem{w22}
Rebecq, H., Gehrig, D., Scaramuzza, D.: Esim: An open event camera simulator. In: Conf. on Robotics Learning (CoRL) (Oct 2018)

\bibitem{m18}
Redmon, J., Farhadi, A.: Yolov3: An incremental improvement (2018), \url{https://arxiv.org/abs/1804.02767}

\bibitem{m9}
Roberto~Vezzani, C.G.A.: University of modena and reggio emilia, \url{https://aimagelab.ing.unimore.it/imagelab/researchActivity.asp?idActivity=055}

\bibitem{w26}
Ryan, C., Elrasad, A., Shariff, W., Lemley, J., Kielty, P., Hurney, P., Corcoran, P.: Real-time multi-task facial analytics with event cameras. IEEE Access  \textbf{11},  76964--76976 (2023). \doi{10.1109/ACCESS.2023.3297500}

\bibitem{m3}
Ryan, C., O’Sullivan, B., Elrasad, A., Cahill, A., Lemley, J., Kielty, P., Posch, C., Perot, E.: Real-time face \& eye tracking and blink detection using event cameras. Neural Networks  \textbf{141},  87--97 (2021). \doi{https://doi.org/10.1016/j.neunet.2021.03.019}, \url{https://www.sciencedirect.com/science/article/pii/S0893608021001076}

\bibitem{m6}
Sefidgar~Dilmaghani, M., Shariff, W., Farooq, M.A., Lemley, J., Corcoran, P.: Optimization of event camera bias settings for a neuromorphic driver monitoring system. IEEE Access  \textbf{12},  32959--32970 (2024). \doi{10.1109/ACCESS.2024.3371487}

\bibitem{m16}
Shahriari, B., Swersky, K., Wang, Z., Adams, R.P., de~Freitas, N.: Taking the human out of the loop: A review of bayesian optimization. Proceedings of the IEEE  \textbf{104}(1),  148--175 (2016)

\bibitem{w25}
Shariff, W., Dilmaghani, M.S., Kielty, P., Moustafa, M., Lemley, J., Corcoran, P.: Event cameras in automotive sensing: A review. IEEE Access  \textbf{12},  51275--51306 (2024). \doi{10.1109/ACCESS.2024.3386032}

\bibitem{m7}
Shijie, L., Yinqiang, Z., Lei, Y., Bin, Z., Xiaowei, L., Jia, P.: Autofocus for event cameras. In: The IEEE/CVF Conference on Computer Vision and Pattern Recognition (CVPR) (2022)

\bibitem{m10}
Singer, S., Nelder, J.: {N}elder-{M}ead algorithm. Scholarpedia  \textbf{4}(7), ~2928 (2009). \doi{10.4249/scholarpedia.2928}, revision \#91557

\bibitem{w2}
Skorka, O., Romanczyk, P.: A review of ieee p2020 noise metrics. Electronic Imaging pp.~1--6 (2022). \doi{10.2352/ei.2022.34.16.avm-109}

\bibitem{w12}
User guide - biasing dynamic sensors. \url{https://gitlab.com/inivation/inivation-docs/-/tree/master/} (2022), accessed: 2022-2-9

\end{thebibliography}
\end{document}